\documentclass{article}
\usepackage{amssymb}
\usepackage{rotating}

\usepackage[english]{babel}

\usepackage[letterpaper,top=2cm,bottom=2cm,left=3cm,right=3cm,marginparwidth=1.75cm]{geometry}
\usepackage{authblk}

\usepackage{amsmath}
\usepackage{graphicx}
\usepackage[colorlinks=true, allcolors=blue]{hyperref}
 \usepackage[square, comma, sort&compress]{natbib}
\usepackage{lineno}

\title{Fourier analysis of the physics of transfer learning for data-driven subgrid-scale models of ocean turbulence}

\author{Moein Darman$^{1}$, Pedram Hassanzadeh$^{2}$, Laure Zanna$^{3}$, and Ashesh Chattopadhyay $^{1}$\thanks{aschatto@ucsc.edu}}
\affil{\normalsize $^1$Department of Applied Mathematics, University of California, Santa Cruz, Santa Cruz}
\affil{\normalsize $^2$Department of Geophysical Sciences, University of Chicago, Chicago}
\affil{\normalsize $^3$Department of Atmospheric and Oceanic Sciences and Mathematics, New York University,  New York}
\begin{document}

\maketitle

\begin{abstract} 
Transfer learning (TL) is a powerful tool for enhancing the performance of neural networks (NNs) in applications such as weather and climate prediction and turbulence modeling. TL enables models to generalize to out-of-distribution data with minimal training data from the new system. In this study, we employ a 9-layer convolutional NN to predict the subgrid forcing in a two-layer ocean quasi-geostrophic system and examine which metrics best describe its performance and generalizability to unseen dynamical regimes. Fourier analysis of the NN kernels reveals that they learn low-pass, Gabor, and high-pass filters, regardless of whether the training data are isotropic or anisotropic. By analyzing the activation spectra, we identify why NNs fail to generalize without TL and how TL can overcome these limitations: the learned weights and biases from one dataset underestimate the out-of-distribution sample spectra as they pass through the network, leading to an underestimation of output spectra. By re-training only one layer with data from the target system, this underestimation is corrected, enabling the NN to produce predictions that match the target spectra. These findings are broadly applicable to data-driven parameterization of dynamical systems.
\end{abstract}




\section{Introduction}

Improving the accuracy of climate, weather, and ocean models requires enhancing their resolution. This necessitates more computational power, which is restricted by limitations. Practical models that run on low spatio-temporal resolution with high predictive accuracy require an accurate representation of subgrid-scale (SGS) processes that occur below the numerical model's grid size. Although these processes occur on such a small spatial and temporal scale that resolving them inside a model is computationally intractable, their effects on the large-scale dynamics is significant. SGS parameterization approximates these processes in models, allowing for more accurate simulations at lower computational costs. The accurate parameterization of SGS processes is essential for improving the accuracy of predictions of these nonlinear, multi-scale, and high-dimensional systems. Consequently, modeling SGS processes has been an active area of research for the past few decades \citep{Dipankar2015, Meneveau2000, Pressel2017, sagaut2013multiscale, Sarlak2015, Schneider2017, Bracco2024,bracco2025machine}.

In the early stages of global climate modeling, \cite{SMAGORINSKY1963} introduced a physics-based SGS model, aiming to parameterize the effects of SGS eddies through a scale-selective dissipative approach, marked by positive eddy viscosity and second-order diffusion. This model and its variants since then, have found applications across a wide array of fields, including weather and climate simulation, combustion, and magnetohydrodynamics, amongst others \citep{ARAKAWA1977, Fox2012, FoxKemper2008, Knaepen2004, Piomelli1999, Piomelli1999, sagaut2005large, Stevens2018, Tan2017}. Despite their utility in ensuring numerical stability for Large Eddy Simulations (LES), these purely diffusive models often fail to accurately represent inter-scale physical processes like energy transfers, notably omitting backscattering \citep{pope2000turbulent}. Backscattering -- the transfer of energy from subgrid to resolved scales -- is critical in various problems involving turbulent fluid flow and has prompted extensive research to incorporate it into physics-based SGS models \citep{Carati1995, Domaradzki1987, Kerr1996, Khani2016, Leslie1979, Mason1992, Mason1992, Shinde2020, Thuburn2013, Zhou1991, Guan2024, Jansen2014, Jansen2019}. Efforts have been made to refine these models for more accurate energy transfer depiction, such as the dynamic approach by \cite{grooms2023backscatter,Germano1991} that allows for negative eddy viscosity to account for backscattering \citep{Grooms2023}. However, these advancements often come with trade-offs in numerical stability, highlighting the ongoing challenge of developing SGS models that precisely capture both forward and backscatter energy transfers. This gap underscores the limitations of physics-based models in SGS parameterization, paving the way for exploring data-driven approaches in subsequent research efforts.

The mathematical relationship between the large-scale dynamics of turbulent flows and the corresponding subgrid-scale forcing is generally nonlinear and poses a significant challenge for researchers. Therefore, neural networks (NNs), as universal approximators \citep{Hornik1989}, are attractive tools to establish such mapping and unveil more hidden knowledge from data, potentially providing better SGS models and even new insights into SGS physics. Nevertheless, some issues exist with relying solely on data-driven models to find the mapping between large-scale and small-scale processes. These models require a large training dataset that contains accurate SGS terms obtained from high-fidelity sources such as high-resolution observations or simulations. However, these sources are usually scarce, and the models only perform well on the dataset on which they were trained. They cannot adapt easily to new datasets, which reduces their effectiveness in predicting dynamics beyond their training dataset. On the other hand, the ability to generalize to newer dynamical regimes is crucial in both SGS modeling and the broader area of machine learning for the physical sciences. For example, NN-based SGS models must perform reliably across diverse climatic conditions to be effectively utilized in projections of global warming \citep{Rasp2018, Larraondo2019} and its impact on extreme weather events, events that can only be estimated from the tails of the probability density function (PDF) of the variables. Additionally, NN-based SGS models are less interpretable, making it difficult to interpret the mappings they create, thereby challenging the trustworthiness and reliability of their predictions.


Given the fact that extrapolation to different climate conditions is an out-of-distribution generalization problem and is challenging for NNs \citep{Krueger2020}, transfer learning (TL) is a flexible and robust framework that enables this \citep{Yosinski2014, Zhuang2021} and can help effective blending of disparate training sets. It involves building a new NN called \textit{Transfer Learned} NN (TLNN) from \textit{Base} NN (BNN), which can achieve a similar level of accuracy for a target system that may have different statistical and dynamical properties compared to the base system. This is accomplished by re-training a few layers from the BNN using a small amount of data (usually orders of magnitude less data than what was used to train the BNN) from the target system. The process can produce a TLNN with comparable out-of-sample accuracy for the target system as the BNN, despite using only a small amount of re-training data.

Numerous studies have been conducted on applying TL to improve NN generalizability for problems involving partial differential equations (PDEs). These investigations predominantly center around Physics-Informed Neural Networks (PINNs) \citep{Chen2021, Desai2021, Gao2022, Goswami2020, Guo2022, Haghighat2021, Hanna2022, Li2021, Mattheakis2021, Xu2023}, where models are fine-tuned or adapted via a physics-based loss function that is specific to the target PDE/ODE system. \cite{Subramanian2023} explores the efficacy of pre-trained machine learning models through TL across a wide range of physics problems. It reveals that fine-tuning pre-trained models reduces the need for extensive downstream training datasets, achieving desired accuracy levels even for out-of-distribution tasks. \cite{Desai2021} developed a framework utilizing TL with PINNs for one-shot inference across ODE and PDEs, demonstrating instant, highly accurate solutions for equations like first- and second-order linear ordinary equations, the Poisson equation and the time-dependent Schrödinger equation, without requiring comprehensive re-training of the network. Additionally, the application of TL to neural operators, defined as operators that learn mappings between two functional spaces from a finite set of input-output pairs (representing coefficients, initial, or boundary conditions as inputs and the PDE solution function as outputs), has garnered significant interest among researchers \citep{Goswami2022, Xu2022, Li2021}. \cite{Goswami2022} presents a TL framework utilizing a deep operator network (DeepONet) to solve nonlinear PDEs in complex geometry other changing dynamics. This framework efficiently addresses task heterogeneity and conditional shifts by fine-tuning particular layers, showcasing rapid learning capabilities despite significant variations between source and target domains. \cite{Xu2022} enhances the stability and long-time prediction accuracy of DeepONet for PDEs by utilizing TL. It involves sequentially updating DeepONets with minimal re-training to track evolution equations over different time frames, demonstrating improved accuracy and reduced training data requirements. The necessity of TL in PINNs underscores the vital importance of comprehending TL mechanisms. This understanding is crucial in enhancing the adaptation and efficacy of models across diverse physical systems, ensuring their broad applicability and performance.



In the context of using TL for improving parameterization generalizability, \cite{chattopadhyay2020data} also showed that TL can improve the generalization skill of data-driven parameterization when they move from one Lorenz 96 system to a more chaotic one. \cite{subel2021data} have shown that TL enables accurate/stable generalization to a flow with 10x higher Reynolds number (Re) for forced Burgers turbulence. \cite{guan2022stable} demonstrated that TL, through re-training the \textit{Convolutional NN} (CNN) using a minimal subset of data from the new flow, achieves accurate and stable LES-CNN predictions for flows at 16× higher Re, and supports higher spatio-temporal resolutions when necessary for achieving stability. \cite{Sun2023} has used NN-based emulators of the Whole Atmosphere Community Climate Model's (WACCM) physics-based gravity wave (GW) parameterizations as a test case. They showed that the accuracy of this NN-based parameterization of GW reduces for a warmer climate (4×CO2). However, it is significantly improved by applying TL, using \(\approx\) 1 \%  data from the warmer climate.

In order to develop parameterization models that are easier to interpret physically, studies aim to utilize ML methods to discover physical equations for the parameterization. This method utilizes a library of pre-defined physical terms and estimate coefficients such that the parameterization can be expressed as a combination of these coefficients and library terms. \cite{Zanna2020} built a fully data-driven, interpretable model employing relevance vector machines (RVM). They used a library of second-order velocity derivatives and their nonlinear combinations to develop a closed-form model for SGS momentum and buoyancy fluxes. While the model showed promising results in \textit{a priori }(offline) evaluations, it revealed instability in \textit{a posteriori} (online) tests when coupled with a low-resolution ocean solver. \cite{karan2024eqdisc} built on the work by \cite{Zanna2020} and used 2D forced homogeneous isotropic turbulence (2D-FHIT) and Rayleigh-B\'enard convection (RBC) test cases to extend the analysis and showed that optimzing on regular mean squared error (MSE) using such pre-defined library terms would provably converge to the 2nd order term in the Taylor series of expansion of the filtering kernel. 

Generally, most of these studies found that, NN-based parameterizations are more accurate offline and if stabilized in online mode, can lead to a more accurate coupled model. However, an important question remains: \textit{what exactly is learned when a neural network is trained on a dataset without predefined assumptions?} It is crucial to identify the specific physical properties of the system that contribute to the model's effectiveness. Recent advancements have been made in understanding what physics learned through the training of CNNs. \cite{subel2022explaining} pioneered the analysis of physics learned from data when applying TL to CNNs for SGS modeling of 2D isotropic turbulence. They utilized a spectral analysis of kernel weights to elucidate how TL adapts to learn new filters, aligning with the spectral differences between base and target systems. \cite{Pahlavan2024} aimed to establish a link between the kernels of NNs and the local and non-local dynamics involved in gravity wave propagation and dissipation, employing Fourier analysis of the CNN's kernels for this purpose.

Despite ongoing efforts to analyze the kernels, pinpointing the exact reasons behind the success of CNN parameterizations remains challenging. It is understood that training a CNN on a given dataset results in the adaptation of convolutional filters to that specific set of data. However, the underlying reasons why particular filters or combinations thereof are effective in parameterization remain unclear—understanding these mechanisms
could reduce training costs and lessen dependence on large volumes of high-fidelity data by enabling more physics-informed initialization of model weights. This paper aims to explore the necessity of TL and its impact on the distribution of kernel weights of CNNs used to parameterize the small-scale dynamics of an anisotropic canonical ocean model. The process of deciding \textit{what-} and \textit{how-to-learn} does not always align with the intended learning outcomes of the model. Therefore, a detailed examination of the kernels offers a pathway to better understand the relationship between the kernels and the physical properties of the system, shedding light on what is actually being learned as opposed to what was intended to be learned. Investigating TL and striving to comprehend its underlying mechanisms offers the dual advantage of fostering a model's generalizability while leveraging this insight to achieve efficient training with fewer samples and enhanced model explainability.

Expanding on the groundwork established by \cite{subel2022explaining}, we incorporate the same analysis for SGS modeling in LES of oceanic flows. Our study broadens the scope to include highly anisotropic and isotropic flow conditions. This approach allows us to explore how CNNs' kernels adapt and learn across varied physical systems, thereby understanding their impact on system performance across various conditions. Our contributions are as follows:

1. We evaluate the CNN-based parameterizations, highlighting offline and online metrics that best assess their out-of-distribution generalization performance.

2. We explain why NNs fail to generalize by linking the spectral characteristics of different flows at different dynamical regimes to the Fourier spectra of activations.

3. We demonstrate that the learnt kernels act as Gabor, low-pass, and high-pass filters and show that their distribution adjusts based on the training data.

This paper is organized as follows. In Section \ref{sec:methods}, we introduce the methodology, including the governing equations of test cases (idealized two-layer quasi-geostrophic model), numerical solver setup, filtering and coarse-graining procedure for data, and CNN and TL employed. Section \ref{sec:results} presents the results. Discussion and summary are in Section \ref{sec:discussions}.

\section{Methods and Data}\label{sec:methods}
    \subsection{Framework Overview}
        The simulations in this study are facilitated by \texttt{pyqg} \citep{pyqg}, a Python library designed for modeling the dynamical behavior of quasi-geostrophic (QG) systems using pseudo-spectral methods. QG systems serve as a suitable approximation for the complex equations governing motion in more realistic ocean models, especially in the limit of high stratification and rotation. They adeptly represent the formation and evolution of ocean mesoscale eddies. Additionally, QG systems offer better computational efficiency compared to ocean models or GCMs, making them vital for the broad scope of this study. We followed the same approach as \citep{ross2023benchmarking}. The methodology detailed in \ref{sec: data collection and preproccesing} repeats their explanations. In section \ref{sec: CNN}, the CNN setup is explained along with the method we use to analyze CNN in spectral space in section \ref{sec: Spectral analysis of CNNs}.

    \subsection{Data Collection and Preprocessing}\label{sec: data collection and preproccesing}
        \subsubsection{Background: Idealized Two-Layer Quasi-Geostrophic Model}
        We utilize a two-layer quasi-geostrophic model provided by \texttt{pyqg}. The prognostic variable of this model is potential vorticity (PV), indicated as \( q_1 \) for the upper layer and \( q_2 \) for the lower layer:
    \begin{equation}\label{PV}
    q_m = \nabla^2 \psi_m + (-1)^m \frac{f^2_0}{g' H_m} \Delta \psi \quad \text{where} \quad m \in \{1, 2\},
    \end{equation}
    where \( \psi_m \) represents the streamfunction corresponding to the depth \( H_m \), \( \Delta \psi = (\psi_1 - \psi_2) \), and \( \nabla = \langle \frac{\partial}{\partial x}, \frac{\partial}{\partial y} \rangle \) denotes the horizontal gradient operator. The zonal and meridional velocities for each layer (\( m \in \{1, 2\} \)) can be derived from the streamfunction as \( u_m = -\frac{\partial \psi_m}{\partial y} \) and \( v_m = \frac{\partial \psi_m}{\partial x} \), respectively. The horizontal velocity vector is expressed as \( \mathbf{u_m} = (u_m, v_m) \). We employ the beta-plane approximation, where the Coriolis acceleration \( f \) varies linearly with latitude (y), described by \( f = f_0 + \beta y \). Here, \( g' \) represents the reduced gravity.
    
    The prognostic equations, solved in the spectral space, are:
    \begin{equation}\label{prognostic equations}
    \frac{\partial \hat{q}_m}{\partial t} = -\hat{J}(\psi_m, q_m) - i k \beta_m \hat{\psi}_m - i k U \hat{q}_m + \delta_{m,2} r_{ek}\kappa^{2}\hat{\psi}^2 + \hat{ssd},
    \end{equation}
    where \( \frac{\partial}{\partial t}\) represents the Eulerian time derivative, \((\hat{\cdot})\) denotes the Fourier transform, and \(\kappa = \sqrt{k^2 + l^2}\) is the radial wavenumber, with \(k\) and \(l\) as the zonal and meridional wavenumbers, respectively. The horizontal Jacobian is defined as \(J(A, B) = A_x B_y - A_y B_x\). The mean PV gradient in each layer is given by \(\beta_m = \beta + (-1)^{m+1} \frac{f_0^2}{g' H_m} \Delta U\), where \(\Delta U = U_1 - U_2\) indicates a fixed mean zonal velocity shear between the two fluid layers. \(U_1\) and \(U_2\) are the mean zonal velocities at the upper and lower levels, respectively. The Dirac delta function \(\delta_{m,2}\) signifies that the bottom drag, with coefficient \(r_{ek}\), is applied solely to the second (bottom) layer. \(\hat{q}\) and \(\hat{\psi}\) are related to each other via
        \begin{equation}
            (\mathbf{M} - \kappa^2 \mathbf{I}) \cdot \begin{bmatrix}
                \hat{\psi}_1 \\
                \hat{\psi}_2
            \end{bmatrix} =
            \begin{bmatrix}
                \hat{q}_1 \\
                \hat{q}_2
            \end{bmatrix},
            \quad \text{where} \quad \mathbf{M} = \begin{bmatrix}
                -\frac{f_0^2}{g' H_1} & \frac{f_0^2}{g' H_1} \\
                \frac{f_0^2}{g' H_2} & -\frac{f_0^2}{g' H_2}
            \end{bmatrix},
        \end{equation}
        where either \(q\) or \(\psi\) can independently identify the state of the system. We will define \(ssd\) below. \\
        \subsubsection{Numerical Solver Setup}\label{sec:Numerical Solver Setup}
        The model is solved pseudospectrally \citep{fox1973pseudospectral} by transforming the velocity field and PV to real space, calculating the Jacobian using real-space PV fluxes, and then transforming back to spectral space. The scale-selective dissipation (\(ssd\)), included as an additive term in Equation \ref{prognostic equations}, is a highly scale selective operator that attenuates the largest \(1/3\) of the spatial wavenumbers of all terms on the right-hand side of Equation \ref{prognostic equations}. Specifically, the operator is an exponential filter, \( F_c(\kappa) \), defined as:
        \begin{equation}\label{filter function}
        F_c(\kappa^*) = 
        \begin{cases} 
        1, & \kappa^* < \kappa_c \\
        e^{-23.6(\kappa^*-\kappa_c)^4}, & \kappa^* \geq \kappa_c 
        \end{cases}
        \end{equation}
        where \( \kappa^* \) is the non-dimensional radial wavenumber and \( \kappa_c = 0.65\pi \), is the cut-off wavenumber. After each time step, \( \hat{q}_m(\kappa^*) \) values are multiplied by \( F_c(\kappa^*) \). Similar to the \( 2/3 \) dealiasing rule \citep{orszag1971elimination}, this filtering scheme reduces aliasing errors in the same range of scales while also providing the numerical dissipation necessary for stable simulations.
        
        We configure the model with a doubly periodic square domain of size \( L = 10^6 \) m, featuring flat topography and a total depth of \( H = H_1 + H_2 \). The model includes a fixed mean zonal velocity shear, \( \Delta U \), with \( U_2 = 0 \). For our different cases, we use different deformation radii \( r_d \), which is the characteristic scale for baroclinic instability and mesoscale turbulence, defined as \( r_d^2 = \frac{g'}{f_0^2} \frac{H_1 H_2}{H} \).
        
        We select the model's grid size, \( \Delta x \), based on the deformation radius. To accurately resolve mesoscale eddies, \( r_d/\Delta x \) must be greater than 2 \citep{hallberg2013using}. For \( r_d = 20,000 \) m, a \( 256 \times 256 \) grid with \( \Delta x_{\text{hires}} = L/256 = 3906.25 \) m yields \( r_d/\Delta x_{\text{hires}} = 5.12 \), ensuring mesoscale turbulence is well-resolved. Conversely, a \( 64 \times 64 \) grid with \( \Delta x_{\text{Lowres}} = L/64 = 15,625 \) m results in \( r_d/\Delta x_{\text{Lowres}} = 1.28 \), which is insufficient to realistically simulate mesoscale eddies. In this lower-resolution setup, parameterization is necessary to account for the unresolved SGS processes. All simulations are conducted with a numerical timestep of \( \Delta t = 1 \) hour.
        
       We consider four different cases categorized into two flow regimes to test our parameterization's generalization ability shown in Fig. \ref{fig: DiffernetCases}: the \textit{eddy configuration}, which leads to the formation of isotropically distributed eddies, and the \textit{jet configuration}, which results in the formation of anisotropic jets. These configurations exemplify the two primary scaling regimes of meridional heat transport \citep{gallet2021quantitative}.
                
        \subsubsection{Predicting Subgrid Forcing}
        We aim to create a mapping from low-resolution velocity (obtained by filtering and coarse-graining the high-resolution velocity fileds) to the subgrid forcing in potential vorticity, \( \Pi_{q_m} \). First, we quantify the subgrid forcing by filtering and coarse-graining high-resolution simulations. This approach assumes the coarsened high-resolution data distribution is similar to low-resolution data for the same data-driven parameterizations to be applicable. We denote the filtering and coarse-graining operator as \( \left(\Bar{\circ} \right) \). The prognostic equation, after filtering, can be expressed as:
        \begin{equation}\label{LES-equation}
            \frac{\partial \hat{\overline{q}}_m}{\partial t} = -\hat{\overline{J}}(\overline{\psi}_m, \overline{q}_m) - i k \beta_m \hat{\overline{\psi}}_m - i k U_m \hat{\overline{q}}_m + \delta_{m,2} r_{ek} \kappa^2 \hat{\overline{\psi}}_2 + \hat{ssd} + \hat{\Pi}_{q_m},
        \end{equation}

        Let \(\partial_t^H\) and \(\partial_t^L\) represent the tendency operators for high- and low-resolution models, respectively. For any given high-resolution \(q\), its subgrid forcing (encompassing nonlinear advection and numerical dissipation( \citep{ross2023benchmarking, Kent2016, PortaMana2014, Shevchenko2021})) is defined as:  
        \begin{equation}
        \Pi_{q_m} = \overline{\partial_t^H q_m} - \partial_t^L \bar{q}_m.
        \end{equation}
        We calculate $\Pi_{q}$ by initializing the high- and low-resolution models with \(q\) and \(\bar{q}\), respectively, advancing both models one step with the same \(\Delta t\), and then subtracting the low-resolution model's tendency from the filtered and coarse-grained tendency of the high-resolution model.
        
        \subsubsection{Filtering and Coarse-graining}
        For data generation, we first coarse-grain and then filter, which are commutative operations for elementwise spectral filtering. Coarse-graining involves reducing the simulation's resolution by a factor of \( K \), specifically by truncating the spatial modes of \( \hat{q} \) to retain only the first \( 1/K \) modes. Spectral filtering typically applies selective decay, diminishing the strength of the highest wavenumbers while retaining the low-wavenumber components after truncation. For the filtering, we apply the Gaussian filter to all remaining modes as follows:
        \begin{equation}\label{GaussianFilter}
        \hat{\Bar{q}}_k = \hat{q}_k * e^{-\kappa^2(2\Delta x_{\text{Lowres}})^2/24}
        \end{equation}
        This is a commonly used filter in SGS modeling \citep{pope2000turbulent,guan2022stable,karan2024eqdisc} where the filter width is chosen to be twice as large as the grid size of the coarse model \citep{lund1997use}.
    \subsection{Convolutional Neural Network (CNN) and Transfer Learning (TL)}\label{sec: CNN}
        Building on our previous work \citep{guan2022stable, guan2022learning, subel2022explaining}, we develop non-local data-driven SGS parameterizations for each case by training a CNN. The CNN takes input \( \mathbf{\Bar{u}_m} = (\Bar{u}_m(x, y), \Bar{v}_m(x, y)) \) and predicts \( \Pi_{q_m}(x, y) \) as the output, where \( m \in \{1, 2\} \) represents the upper and lower levels. These CNNs consist of 9 sequential convolution layers, with 7 hidden layers each containing \( 64^2 \) kernels of size \( 5 \times 5 \). The outputs of a convolutional layer, called activations, are denoted for channel \( j \) of layer \( \ell \) as \( g_{\ell}^{j} \in \mathbb{R}^{N_{\text{Lowres}} \times N_{\text{Lowres}}} \), and the activation equation is:
        \begin{equation}\label{eq:activations}
                g_\ell^j(\mathbf{u}) = \sigma\left(\sum_\beta \left( W_\ell^{\beta,j}\circledast g_{\ell-1}^\beta(\mathbf{u})\right)+b_\ell^j\right).
        \end{equation}
        Note that \( N_{\text{Lowres}} = 64 \) for all cases. Here, \(\circledast\) represents spatial convolution and \( \sigma(\circ) = \max(0, \circ) \) is the ReLU activation function (absent in the linear output layer, \( \ell = 9 \)). \( W^{\beta,j}_{\ell} \in \mathbb{R}^{5 \times 5} \) is the weight matrix of a convolution kernel, and \( b^j_{\ell} \in \mathbb{R}^{64 \times 64} \) is the regression bias, a constant matrix. For all layers, \( \beta \in \{1,2,\ldots,64\} \) and \( j \in \{1,2,\ldots,64\} \) except in the input layer (\( \ell = 1 \)), where \( \beta \in \{1,2,3,4\} \), and in the output layer (\( \ell = 9 \)), where \( j \in \{1,2\} \) as the output has two channels. The kernels' weights and biases are the trainable parameters of the NN, collectively referred to as \( \theta \in \mathbb{R}^p \). Note that \( g_{\text{in}} = g_0 = \mathbf{\Bar{u}_m} \) and \( g_{\text{out}} = g_{9} = \Pi_{q_m} \).
        
        We refer to CNNs that are trained from scratch on each \(\text{Case}_i\) and tested on \(\text{Case}_j\) as \(\text{BNN}^{i,j}\). All the trainable parameters \(\theta\) are \textit{randomly} initialized. BNNs are trained with 40,000 samples from each case. Training is conducted for 100 epochs with a starting learning rate of \(l_r = 10^{-3}\) and a scheduler that reduces the learning rate by factor of 10 when metrics stop improving to avoid overfitting on each dataset. We used a batch size of 8 and the ADAM optimizer \citep{kingma2014adam}. The mean-squared error (MSE) was used as the loss function, and the model performance was evaluated using the metrics introduced in section \ref{sec:offline metrics}.
        
\begin{figure}[h]
            \centering
            \makebox[\textwidth]{
                \includegraphics[width=\textwidth, clip, trim=0cm 19.4cm 0cm 0cm]{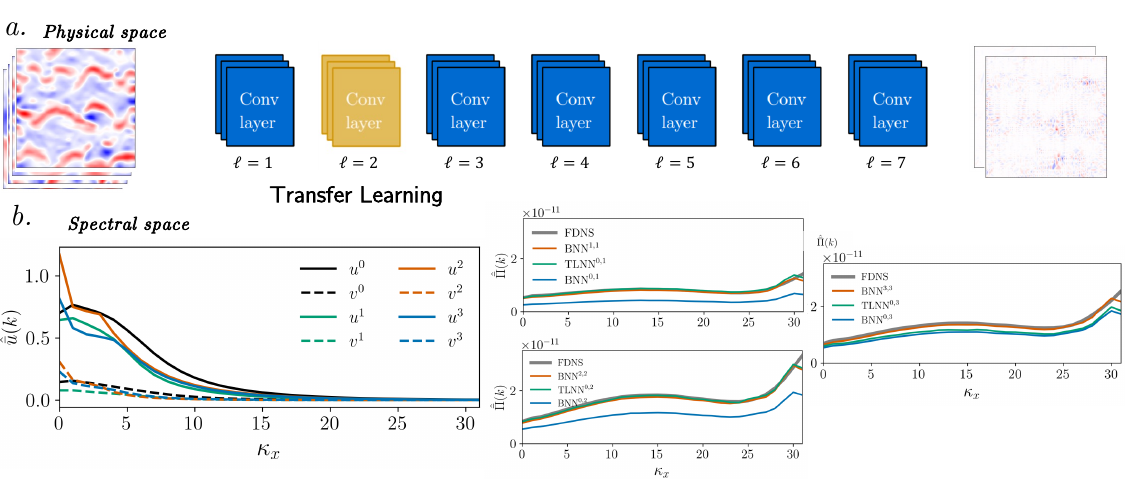}
            }
            \caption{\textit{Row a} displays the schematic of the CNN and inputs and outputs in physical space. Each TLNN is initialized with the weights of $\text{BNN}^{0}$, and only the first hidden layer ($\ell = 2$) is re-trained using a smaller percentage of data. The inputs are the meridional and zonal velocities of the upper and lower levels, and the output is the subgrid forcing for each level.
                \textit{Row b} shows the inputs and outputs of the CNN in spectral space, with the spectrum meridionally averaged}
            \label{fig:schematic}
        \end{figure}

        For TL, we initialize the weights and biases of \(\text{BNN}^{0}\) and retrain only the first hidden layer (\(\ell = 2\)) with different percentages of data from the target case, using the same setup for learning as the BNNs. The models utilizing transfer learning from \(\text{Case}_i\) to \(\text{Case}_j\) are referred to as \(\text{TLNN}^{i,j}\). This approach allows us to use less data to achieve good performance on the target case. The schematic of the CNN used in parameterization, its architecture, and its input/output in physical space are shown in Fig. \ref{fig:schematic}. 

    \subsection{Spectral Analysis of CNNs}\label{sec: Spectral analysis of CNNs}
        The Fourier transform operator $\mathcal{F}$ is defined as
        \begin{equation}
        \label{eq:fft_normal}
            \hat{(\circ)} \equiv \mathcal{F} \left(\circ  \right), \quad \mathcal{F}: ~ \mathbb{R}^{64\times 64} \longmapsto \mathbb{C}^{64\times 64}.
        \end{equation}    
        To express convolution in the spectral domain, we first note that each kernel \( W_\ell^{\beta,j}\in \mathbb{R}^{5\times 5} \) can be extended to the full domain of the input by zero-padding, a common practice for faster training \citep{mathieu2013fast}, resulting in \( \widetilde{W}_\ell^{\beta,j} \in \mathbb{R}^{64\times 64} \). Using the convolution theorem, we then obtain
        \begin{equation}
        \label{eq:fft_weights}
             W_\ell^{\beta,j}\circledast g_{\ell-1}^\beta = \mathcal{F}^{-1}\left(\hat{\widetilde{W}}_\ell^{\beta,j}\odot \hat{g}_{\ell-1}^\beta\right),
        \end{equation}
        where $\odot$ is element-wise multiplication. \\
        
        Next, we define linear activation $h_\ell^j$, which contains all the linear operations in Eq.~(\ref{eq:activations}):
        \begin{equation}
            h_\ell^j = \sum_\beta \left( W_\ell^{\beta,j}\circledast g_{\ell-1}^\beta\right)+b_\ell^j.
            \label{eq:act_linear}
        \end{equation}
        Even though Eq.~(\ref{eq:activations}) is nonlinear due to the ReLU function, its Fourier transform can still be derived analytically. By applying Eqs.~(\ref{eq:fft_weights}) and \eqref{eq:act_linear} and utilizing the linearity property of the Fourier transform, we obtain
        \begin{equation}
            \hat{g}_{\ell}^j = \sum_{\alpha} \left(e^{-i(k_{x}x_{\alpha}+k_{y}y_{\alpha})}\right) \circledast \hat{h}_{\ell}^j = \sum_{\alpha} \left(e^{-i(k_{x}x_{\alpha}+k_{y}y_{\alpha})}\right) \circledast \left\{\sum_\beta  \left(\hat{\widetilde{W}}_\ell^{\beta,j}\odot \hat{g}_{\ell-1}^\beta\right)+\hat{b}_\ell^j \right\},
            \label{eq:gspectra}
        \end{equation}
        where \((x_\alpha, y_\alpha) \in \left\{(x, y)~|~h_\ell^j(x, y) > 0 \right\}\) and \(i = \sqrt{-1}\). The sum over \(\alpha\) arises from the ReLU function, involving grid points where \(h_\ell^j > 0\). Note that \(e^{-i(k_{x}x_{\alpha}+k_{y}y_{\alpha})}\) represents the Fourier transform of the Heaviside function at \((x_\alpha, y_\alpha)\). Also, since \({b}_\ell^j\) is a constant matrix, \(\hat{b}_\ell^j\) is non-zero only at \(k_x = k_y = 0\) (and is real).
        
        Equation~\eqref{eq:gspectra} indicates that the spectrum of $\hat{g}_{\ell}^j$ is influenced by the spectrum of $\hat{g}_{\ell-1}^j$, the spectra of the weights $\hat{\widetilde{W}}_\ell^{\beta,j}$ (and constant biases $\hat{b}_\ell^j$), and the regions in the physical space where $h_{\ell}^j>0$. With TL, updating the weights and biases changes both their spectra and the regions where $h_{\ell}^j>0$.
        
        To understand how TL enables the TLNN to capture the physics of different cases, we perform extensive analysis on how the weights ($\hat{\widetilde{W}}_\ell^{\beta,j}$) and the spectrum of all layers ($\hat{g}_{\ell}^\beta$) change during TL. This allows us to explain how the distribution of kernels affects the shape of the spectrum and how these spectral changes propagate through the layers to the final layer where the loss function is applied. This process highlights the adaptation of weights to the physics and demonstrates how TL can adjust the weights to be consistent with the physics of the target system.

    \subsection{Offline metrics}\label{sec:offline metrics}
    We use several metrics to  evaluate the model's \textit{a priori} performance. The first metric is root mean square error, defined in Eq.~(\ref{eq: RMSE}), which assesses the optimality of the mapping between input and output. Here, \( N \) is the number of test samples, and \(\left\| \cdot \right\|_2\) represents the L2 norm. This ensures the model prediction and true output are close in the L2 space.
        \begin{equation}\label{eq: RMSE}
        \mathrm{RMSE}_{\Pi_{q_m}} = \sqrt{\frac{1}{N} \sum_{i=1}^{N} \left\| \bar{\Pi}_{\mathbf{q}_{m}i}^\text{pred} - 
        \bar{\Pi}_{\mathbf{q}_{m}i}^\text{true} \right\|_2^2 \Bigg/ \frac{1}{N} \sum_{i=1}^{N} \left\| \bar{\Pi}_{\mathbf{q}_{m}i}^\text{true} \right\|_2^2}.
        \end{equation}
    The correlation coefficients (CC), averaged over \( N \) test samples, assess the CNN model's ability to capture the structure of the true data
    \begin{equation}
        CC_{\Pi_{q_m}} = \frac{
    \langle \left( \bar{\Pi}_{q_m}^{pred} - \langle \bar{\Pi}_{q_m}^{pred} \rangle \right)
    \left( \bar{\Pi}_{q_m}^{true} - \langle \bar{\Pi}_{q_m}^{true} \rangle \right) \rangle
}{
    \sqrt{
        \langle\left( \bar{\Pi}_{q_m}^{pred} - \langle \bar{\Pi}_{q_m}^{pred} \rangle \right)^2
    \rangle}
    \sqrt{
        \langle\left( \bar{\Pi}_{q_m}^{true} - \langle \bar{\Pi}_{q_m}^{true} \rangle \right)^2
        \rangle
    }
},
    \label{eq: cc}
    \end{equation}
    where \(\langle \cdot \rangle\) represents domain averaging.  
    
    We evaluate models performance by calculating the RMSE between the predicted output spectrum and the true output spectrum 
        \begin{equation}
        \text{Spectrum RMSE} = \frac{1}{N_{k_x}} \sum_{k_x} \left| \frac{\hat{\bar{\Pi}}_{q_m}^\text{pred}(k_x) - \hat{\bar{\Pi}}_{q_m}^\text{true}(k_x)}{\hat{\bar{\Pi}}_{q_m}^\text{true}(k_x)} \right|.
        \end{equation}
    Spectrum RMSE is essential for ensuring that the model not only predicts accurate values but also preserves the spectral characteristics of the data.

\section{Results}\label{sec:results}
    In this section, we present and discuss the results of our study. We start by examining the distinct physical and spectral characteristics of different cases. Subsequently, we focus on the generalization capabilities of BNN and the effectiveness of TL to enhance these capabilities both \textit{a priori} and \textit{a posteriori}. Furthermore, we emphasize the significance of spectral RMSE in assessing the model's generalization performance. Additionally, we examine the CNNs in spectral space to observe the impact of retraining one layer on subsequent hidden-layer activation spectra. This section also explains the role of weights as spectral filters and shows and quantifies how kernel distributions optimally adjust in response to changes in training data.
    
    \begin{figure}[h]
    \centering
    \makebox[\textwidth]{%
        \includegraphics[width=\textwidth, clip, trim=0cm 10.4cm 0cm 0.8cm]{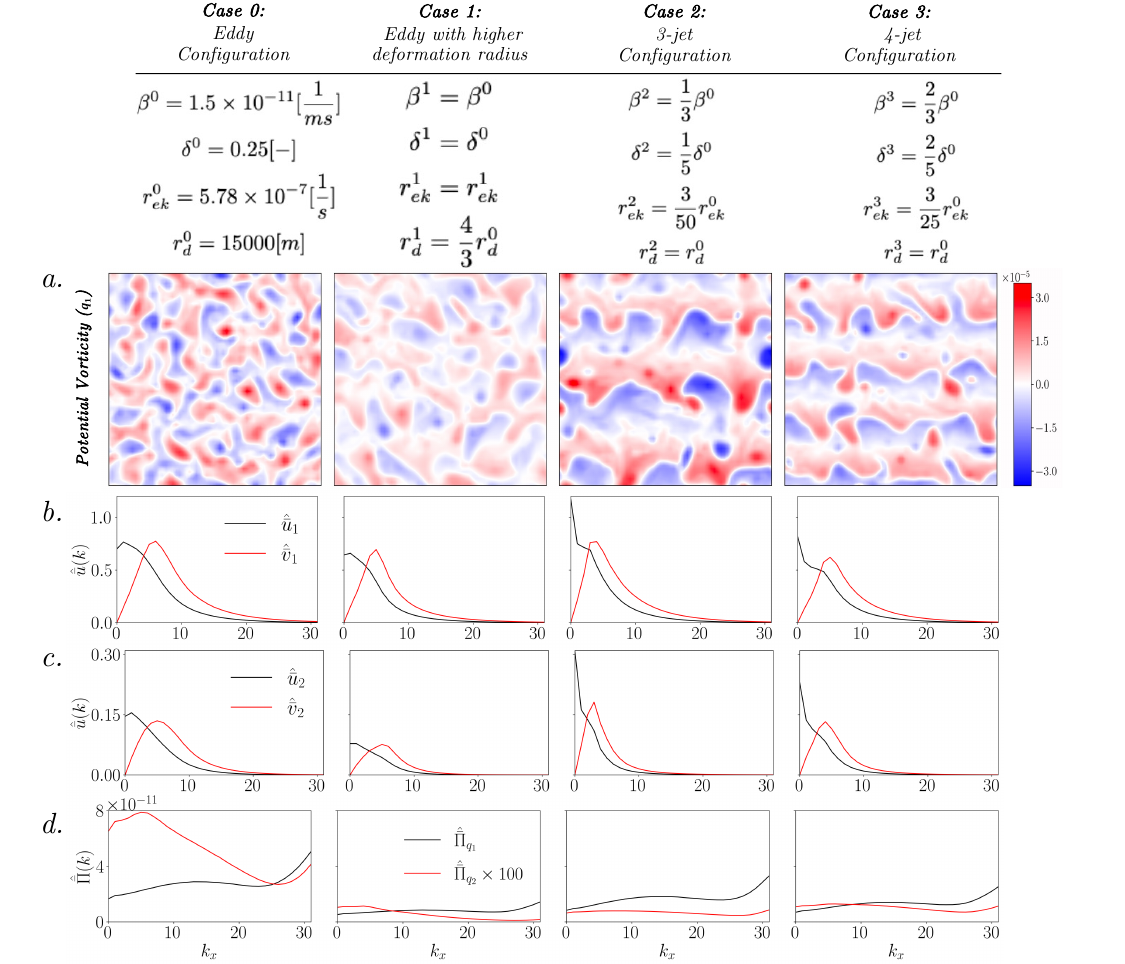}
    }
    \caption{
    Comparative analysis of the base system and three target configurations using 10 years of simulation data. 
    \textit{Row a:} Snapshots of potential vorticity showcasing the spatial distribution of eddies in each system—eddies display a roughly isotropic structure, while jets exhibit more organized zonal alignment. 
    \textit{Row b:} Meridionally averaged spectra of velocity profiles at the upper level. 
    \textit{Row c:} Meridionally averaged spectra of velocity profiles at the lower level. 
    \textit{Row d:} Meridionally averaged spectra of subgrid forcing at both levels. 
    In all spectral panels, \(k_x\) represents the zonal wavenumber, and spectra are averaged over the meridional direction}
    \label{fig: DiffernetCases}
    \end{figure}
    \newpage

    \subsection{Overcoming BNN's Generalization Limits: How TL Bridges the Gap}

    Before explaining the physics behind TL, we need to quantify the improvement in parameterization performance due to TL. Here, we discuss the ways TL improves parameterization and metrics that can best assess the model's generalization. By examining Fig. \ref{fig: DiffernetCases} (a), we observe that adjusting the parameters of Eq.~(\ref{prognostic equations}) leads to different physical characteristics. When examining the potential vorticity snapshots of the upper level ($q_1$) shown in Fig. \ref{fig: DiffernetCases} (a), isotropic eddies are formed for Case 0. The same system's behavior with larger eddies is observed by increasing the deformation radius (Case 1). Altering $\beta$, $\delta$, and $r_{ek}$ leads to creating highly anisotropic jets with varying numbers (Case 2 $\&$ 3). Different physical characteristics are reflected in different spectra, meaning that cases differ in both large and small scales. Figures \ref{fig: DiffernetCases} (b, c, and d) illustrate the distinct physics governing each system, resulting in differences in velocity and subgrid forcing spectra.
    
\begin{figure}[h]
        \centering
        \makebox[\textwidth]{
            \includegraphics[width=\textwidth, clip, trim=0cm 9.5cm 0cm 0cm]{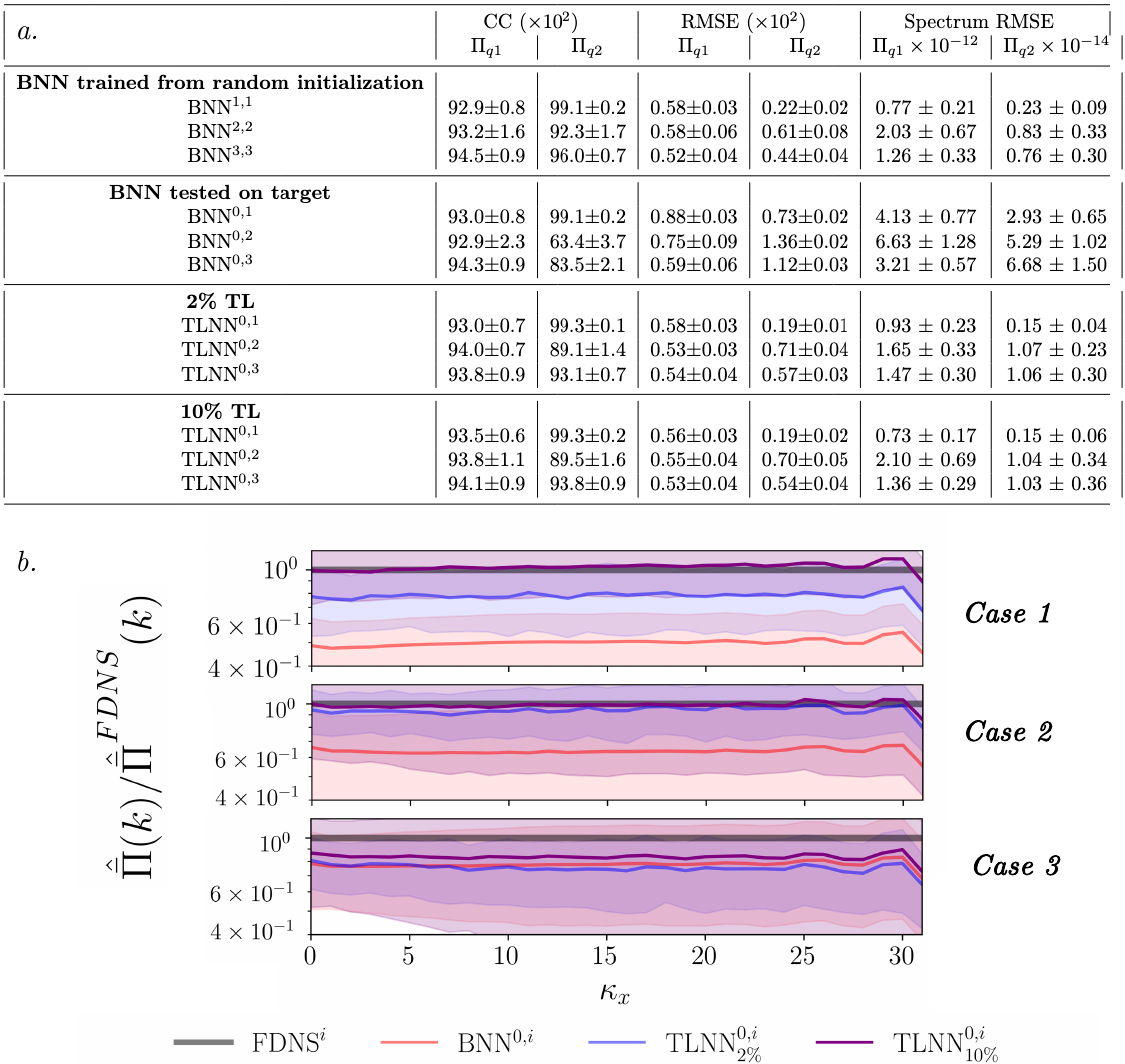}
        }
        \caption{
        \textit{A priori} evaluation of CNN parameterization. 
        \textit{Panel a}: CC, RMSE, and spectrum RMSE for upper and lower levels, comparing $\mathrm{BNN}^{i,i}$, $\mathrm{BNN}^{0,i}$, and $\mathrm{TLNN}^{0,i}$ with different re-training data percentages across three target cases. 
        \textit{Panel b}: Ratio of output spectrum to FDNS spectrum for each case}
        \label{fig:OfflineResults}
    \end{figure}

    Figure \ref{fig:OfflineResults} (a) shows that, when provided with sufficient data, the $\text{BNN}^{i,i}$ performs best, within uncertainty, across all metrics. However, when tested on different cases, $\text{BNN}^{0,i}$ struggles to generalize to other systems with different spectral properties, leading to sub-optimal performance. In particular, $\text{BNN}^{0,i}$ fails to generalize across all metrics, except for the correlation coefficient in the upper layer, which remains comparable to $\text{BNN}^{i,i}$. Furthermore, there is minimal change in the CC of the subgrid forcing at the upper level when more data is integrated into TL. However, the RMSE consistently decreases as more data is included in TL. Spectrum RMSE offers more valuable insights into the model's generalization capabilities. It consistently worsens at both upper and lower levels when the model is extrapolated and improves at both levels when TL is applied with 2\% and 10\% of target system data. These results demonstrate that TL enhances the generalization capability of $\text{BNN}^{0}$ using only a small fraction of the data needed to train a new $\text{BNN}^{i}$ from scratch.
    
    Figure \ref{fig:OfflineResults} (b) shows the ratio of the output spectrum to the FDNS spectrum for each case. $\text{BNN}^{0,i}$ exhibits the largest gap relative to FDNS, while TL progressively closes this gap as more re-training data is used. This further confirms that TL helps align the spectral characteristics of the learned subgrid forcing with that of the true system.
    
\begin{figure}[h]
        \centering
        \makebox[\textwidth]{
            \includegraphics[width=\textwidth, clip, trim=0cm 13.5cm 0cm 0cm]{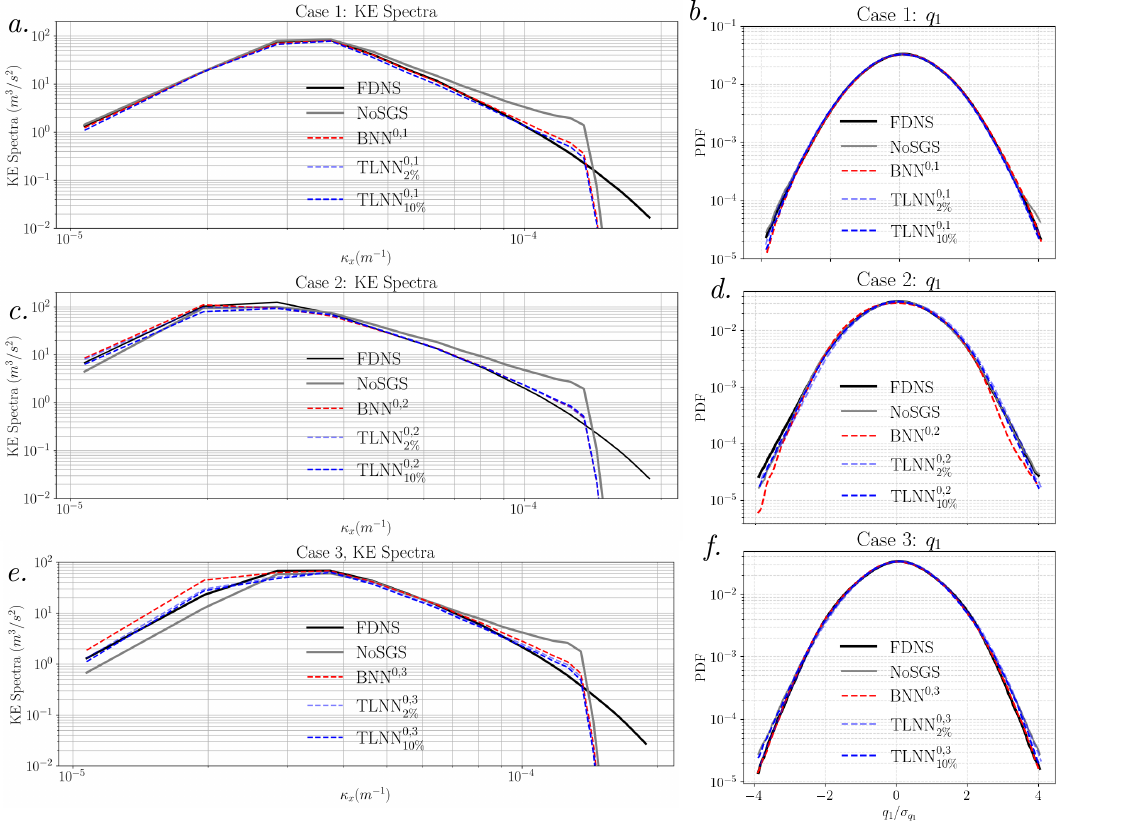}
        }
        \caption{
        \textit{A posteriori} evaluation of CNN parameterization. 
        \textit{Panels a, c, and e}: Kinetic energy spectra from 10-year simulations using $\text{BNN}^{0,i}$ and $\text{TLNN}^{0,i}$ across different cases. 
        \textit{Panels b, d, and f}: PDFs of potential vorticity at the upper level for the same simulations}
        
        \label{fig:OnlineResults}
    \end{figure}

    As shown in Fig. \ref{fig:OnlineResults}, the TLNN can rectify mismatches in the kinetic energy (KE) spectra wherever there is room for improvement. Specifically, Fig. \ref{fig:OnlineResults} (a) shows that $\text{TLNN}^{0,i}$ improves the KE spectra at high wavenumbers compared to $\text{BNN}^{0,i}$, with both parameterizations performing better than the simulation without any SGS parameterization. In contrast, Fig. \ref{fig:OnlineResults} (c) shows that all parameterizations perform equally well in that case, and all surpass the no-parameterization baseline. Fig. \ref{fig:OnlineResults} (e) demonstrates that $\text{TLNN}^{0,i}$ better captures the KE spectra at both low and high wavenumbers compared to $\text{BNN}^{0,i}$.
    
    The scale-selective dissipation (ssd) introduced in Section \ref{sec:Numerical Solver Setup} can obscure the positive effects of TL in improving online results. This is evident in the PDFs of $q_1$ in Fig. \ref{fig:OnlineResults} (b, d, f), where even the baseline simulation without SGS parameterization captures the mean flow well. Nonetheless, TL proves beneficial wherever there is room for improvement. For example, Fig. \ref{fig:OnlineResults} (d) shows that $\text{TLNN}^{0,i}$ improves the PDF tails in Case 2, highlighting its potential to fine-tune models efficiently for extreme events with limited re-training data across different dynamical regimes.

        \subsection{Why Does BNN Fail to Generalize? How Does TL Solve Generalization Issues?}\label{sec:The Physics Behind TL}

        Analyzing the CNN in the spectral space provides a clearer distinction between different physical systems \citep{subel2022explaining, pahlavan2024explainable}. When $\text{BNN}^{0}$ is applied to out-of-distribution inputs, it underestimates the channel-averaged, meridionally averaged activation spectrum relative to $\text{BNN}^{0,0}$, whose weights are tuned for in-distribution data. This underestimation begins in early layers and propagates through the network, ultimately resulting in a mismatch with the FDNS output spectrum.
        
\begin{sidewaysfigure}
        \centering
        \includegraphics[width=\textheight, trim=0cm 17cm 0cm 0cm, clip]{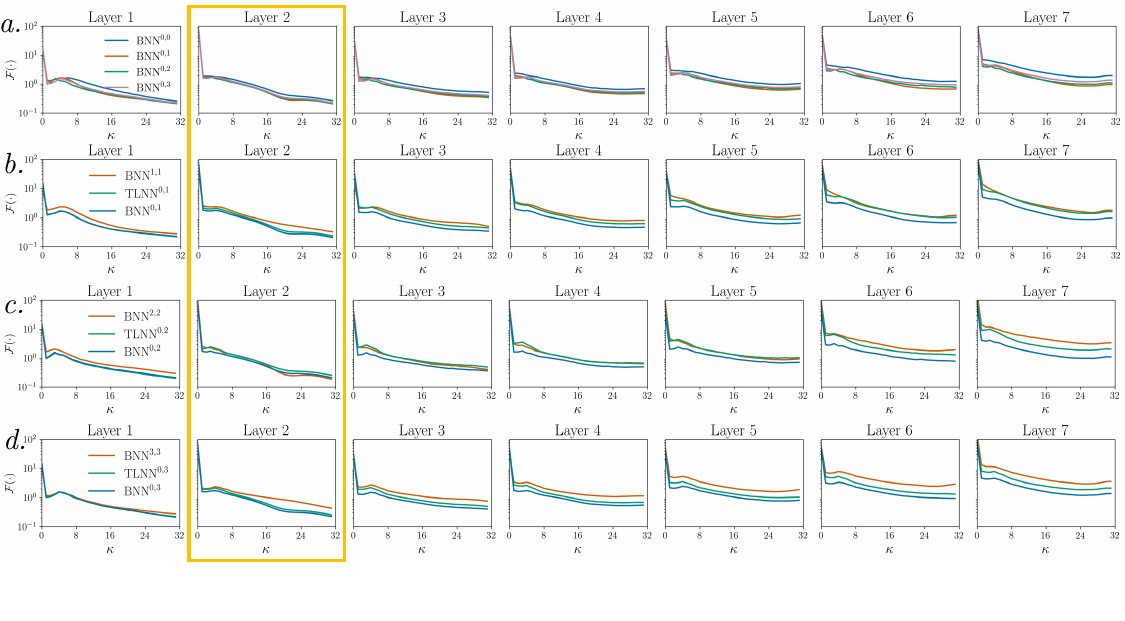}
        \caption{Channel-averaged, meridionally averaged spectra of hidden layer activations across different networks. 
            \textit{Rows a to d} show spectra of hidden layers for different models and cases.}
        \label{fig:CNN}
        \end{sidewaysfigure}

        Figure~\ref{fig:CNN} illustrates this mechanism. \textit{Row a} demonstrates that, when $\text{BNN}^{0,i}$ is tested with out-of-distribution data, underestimation of the spectral content is evident across all layers compared to $\text{BNN}^{0,0}$. This leads to a mismatch in the output spectrum relative to FDNS.
        
        However, applying TL by re-training only the second hidden layer in $\text{TLNN}^{0,i}$ helps close this gap. As seen in \textit{Rows b to d} of Fig.~\ref{fig:CNN}, this localized re-training causes an upshift in the spectrum at layer 2. While this adjustment may be subtle at $\ell = 2$, it propagates through subsequent layers, leading to an output spectrum that better matches the FDNS. The behavior of $\text{TLNN}^{0,i}$ becomes more similar to that of $\text{BNN}^{i,i}$, which serves as the ideal case for each target dataset.
        
        Figure~\ref{fig:CNN} thus highlights how targeted re-training of just one layer can realign the CNN’s internal spectral response with the physics of the new system, enabling generalization across dynamical regimes.

        \begin{figure}[p]
        \centering
        \includegraphics[width=\textwidth, clip, trim=0cm 1.65cm 0cm 0.5cm]{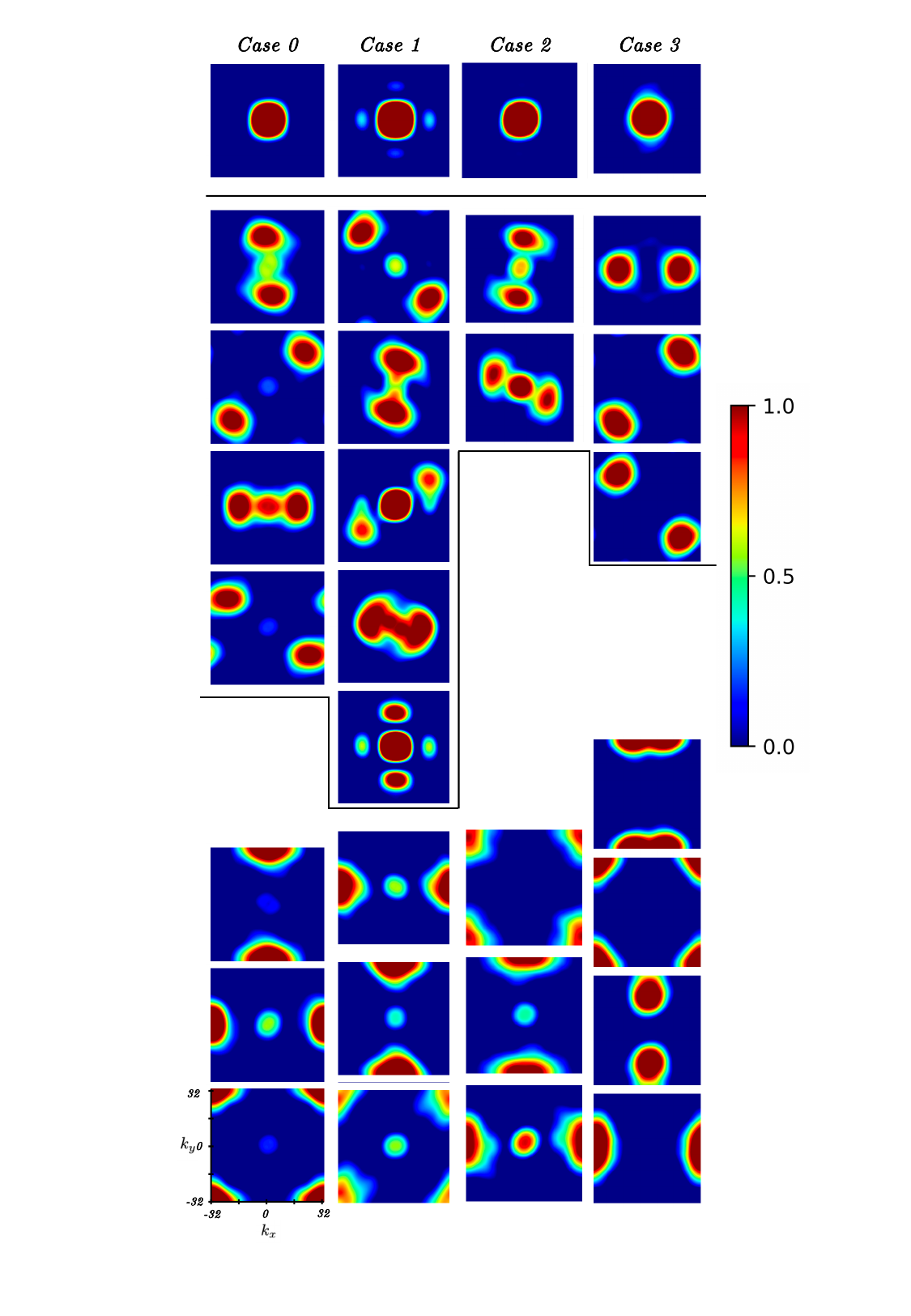}
        \caption{
        Cluster centers of filter spectra obtained by applying the \(k\)-means algorithm to the \(64^2\) padded weight matrices $\left| \hat{\widetilde{W}}_\ell^{\beta,j} \right|$ from layer 2 of \(\text{BNN}^{0}\) and \(\text{TLNN}^{i}\). 
        The number of cluster centers varies in each case as we increase the number of cluster centers until qualitatively similar patterns are observed}
        \label{fig:ClusterCenters}
        \end{figure}

        \subsection{Spectral Filters in Action: The Role of Weights in Generalization}
        
        The weights within a CNN act as spectral filters that influence how information at different spatial scales is processed through the network. To better understand what controls the spectral content of activations $g_{\ell}$, we analyze the convolutional kernels in Fourier space—a powerful approach widely used in understanding the physics of turbulence.
        
        We examine all \(64^2\) kernels in the second convolutional layer. Since direct visualization is impractical, we apply the \(k\)-means clustering algorithm to identify representative cluster centers. As shown in Fig.~\ref{fig:ClusterCenters}, the cluster centers across all four cases consistently represent combinations of low-pass, Gabor, and high-pass filters. This behavior is largely independent of whether the training data are isotropic or anisotropic, and does not, on its own, distinguish between different physical systems. Nevertheless, the spectral magnitudes of the weights exhibit structured peaks at specific wavenumber pairs \((k_x, k_y)\).
        
        To analyze how the distribution of kernel weights adapts during transfer learning, we identify the wavenumber pair \((k_x, k_y)\) where the absolute value of each Fourier-transformed, padded kernel \(\left| \hat{\widetilde{W}}_{2}^{\beta,j} \right|\) reaches its maximum. This gives a spectral "footprint" of each filter’s dominant response. We focus specifically on layer 2—the only layer re-trained during TL—while keeping all other layers frozen (as discussed in Section~\ref{sec:The Physics Behind TL}). This setup allows us to isolate and quantify how the second layer adjusts when adapting to out-of-distribution samples.
        
\begin{figure}[h]
            \centering
            \makebox[\textwidth]{
                \includegraphics[width=\textwidth, clip, trim=0cm 14cm 0cm 0cm]{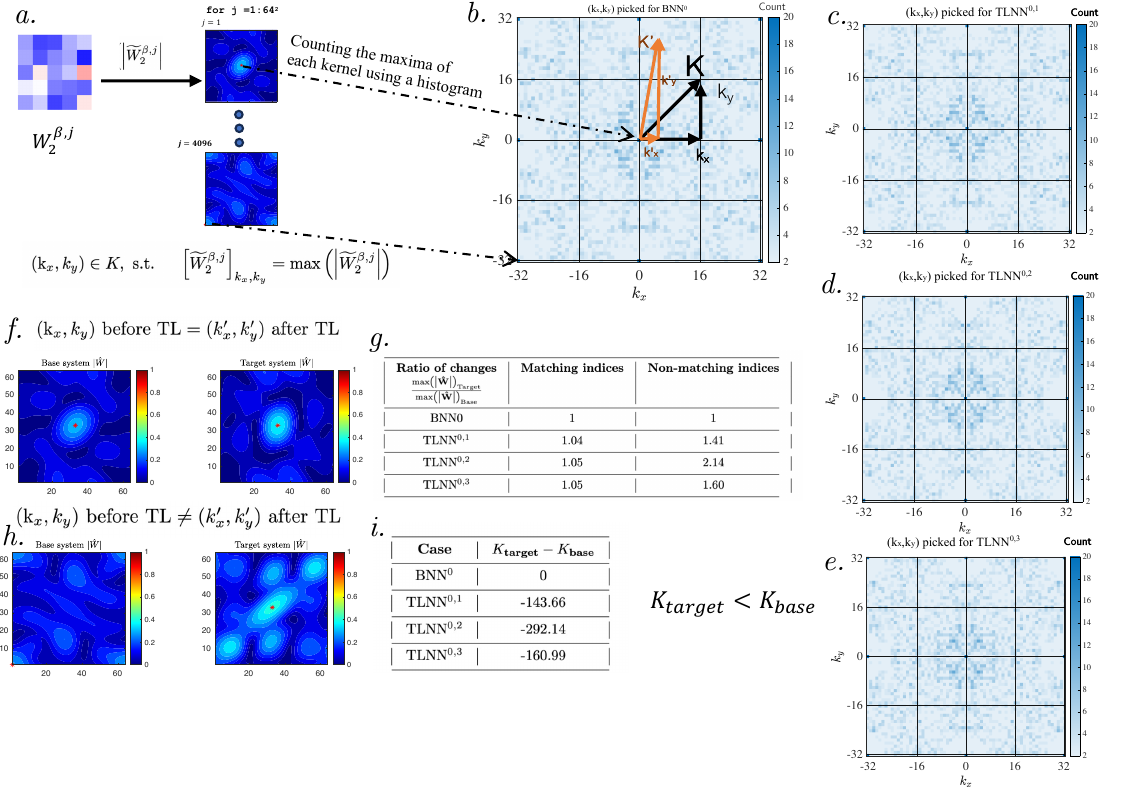}
            }
            \caption{
            Schematic and analysis of kernel changes in spectral space. 
            \textit{Panel a}: Identification of the $64^2$ wavenumber pairs corresponding to the maxima of the padded and Fourier-transformed layer-2 kernels $\left| \hat{\widetilde{W}}_\ell^{\beta,j} \right|$. 
            \textit{Panels b–e}: Histograms of these wavenumber pairs for $\text{BNN}^{0}$ and \(\text{TLNN}^{0,i}\). 
            \textit{Panel f}: Wavenumber pairs whose maxima locations do not change after TL. 
            \textit{Panel g}: Change in the mean amplitude ratio at the maxima. 
            \textit{Panel h}: Wavenumber pairs whose maxima locations shift after TL. 
            \textit{Panel i}: Comparison of change in radial wavenumber between base and target systems}
            \label{fig:TLexplanation}
        \end{figure}

        Histograms of these dominant wavenumber pairs are shown in Fig.~\ref{fig:TLexplanation} (b–e) for both $\text{BNN}^{0}$ and $\text{TLNN}^{0,i}$. Most kernels behave as low-pass or high-pass filters, while Gabor filters appear less frequently and without a dominant orientation. The saturation in the center and corners of each histogram indicates a higher concentration of low-pass and high-pass filters. 
        
        To further analyze how these maxima change under TL, we divide the kernels into two categories:
        
        1. Unchanged maxima locations: For filters whose dominant wavenumber remains the same after TL (Fig.~\ref{fig:TLexplanation} (f)), we compute the mean ratio of spectral amplitude at the maximum before and after re-training. As shown in Fig.~\ref{fig:TLexplanation} (g), this amplitude increases consistently, aligning with the upshift in the activation spectra observed in Fig.~\ref{fig:CNN} (b–d).
        
        2. Shifted maxima locations: For kernels whose dominant wavenumber changes after TL (Fig.~\ref{fig:TLexplanation} (h)), we compute the radial wavenumber \( \kappa = \sqrt{k_x^2 + k_y^2} \) before and after TL to determine how these shifts affect the scale preference of the filters. Fig.~\ref{fig:TLexplanation} (i) shows that TL shifts many of these maxima toward lower wavenumbers, indicating a stronger focus on large-scale features. This behavior is consistent with the decaying nature of the activation spectra and supports the idea that adapting to new dynamics requires capturing the dominant large-scale structure.

    \section{Discussion}\label{sec:discussions}
    
    In Section~\ref{sec:results}, we introduced a non-intrusive approach to explain how kernels adapt during TL with minimal data. A detailed analysis of the weights in BNNs and TLNNs reveals that the learned kernels function as low-pass, Gabor, and high-pass filters, regardless of whether the training data are isotropic or anisotropic. This study presents the first comprehensive effort to relate the distribution of these spectral filters to the spectral characteristics of isotropic and anisotropic flows in the context of SGS modeling.
    
    The methodology proposed here advances our understanding of what is learned during TL from physical data, particularly in high-dimensional systems. Rather than viewing TL purely as an optimization task, this study aims to uncover the underlying mechanisms that govern learning from limited data. For instance, our findings suggest that spectral bias~\citep{chattopadhyay2023long,chattopadhyay2024oceannet,gray2024long,lupin2025simultaneous} in NNs—where filters underrepresent high-wavenumber content—can be partially explained by how weights evolve during training.
    
    While this analysis assumes that filters have a single global maximum, some kernels exhibit multiple significant local maxima that pass activation at different scales. Although this assumption simplifies the analysis, more detailed studies could provide a more precise understanding of what is learned during TL. Additionally, while clustering offers a starting point for explaining the learned kernels as spectral filters, more is needed to fully explain how these kernels process information at different scales. Furthermore, using radial wavenumber to analyze kernel spectra distribution may overlook directionality, necessitating further research to address this limitation. It is also important to note the role of numerical dissipation in the solver, which may obscure the improvements brought by TL.
    
    Although our findings are likely specific to the test cases, network architecture, and SGS parameterization studied here, the analysis methods are broadly applicable. They can be extended to a wide range of base–target system combinations and applications, including data-driven forecasting and training set blending. The comprehensive analysis introduced here holds potential for many multi-scale dynamical systems applications.
    
    In conclusion, we examined the performance of CNN-based parameterizations for quasi-geostrophic turbulence, focusing on various offline metrics to determine which best explains generalization performance. By analyzing CNNs in the spectral space, we uncovered why generalization fails in these systems, specifically by examining activation spectra and their propagation through subsequent layers. To understand how TL can improve this sub-optimal performance, we investigated how kernel distributions adapt to the training dataset. Identifying the quantity and intensity of these filters before and after TL allowed us to explain how TL effectively bridges the generalization gap.


\paragraph{Acknowledgments}
We express our gratitude to Karan Jakhar, Rambod Mojgani, and Hamid A. Pahlavan for their valuable feedback during this project.

\paragraph{Funding Statement}
PH was supported by ONR award N000142012722 and NSF grant AGS-2046309. AC and MD were supported by NSF grant no. 2425667 and computational support from NSF ACCESS MTH240019 and NCAR CISL UCSC0008 and UCSC0009.

\paragraph{Data Availability Statement}
The code used for the analysis in this project is available on GitHub at the following link: 
\url{https://github.com/moeindarman77/TransferLearning-QG}

\paragraph{Ethical Standards}
The research meets all ethical guidelines, including adherence to the legal requirements of the study country.

\paragraph{Author Contributions}
P.H., A.C., and L.Z. conceptualized the research. M.D. conducted the research. A.C. contributed to some of the early versions of the computational codes. A.C. and M.D. wrote the draft. All authors analyzed the results and edited the manuscript. 

\paragraph{Competing Interests} 
The authors declare that they have no competing interests. 


\bibliographystyle{abbrvnat}
\bibliography{References}

\begin{thebibliography}{78}
\providecommand{\natexlab}[1]{#1}
\providecommand{\url}[1]{\texttt{#1}}
\expandafter\ifx\csname urlstyle\endcsname\relax
  \providecommand{\doi}[1]{doi: #1}\else
  \providecommand{\doi}{doi: \begingroup \urlstyle{rm}\Url}\fi

\bibitem[Abernathey et~al.(2022)Abernathey, {Rochanotes}, Ross, Jansen, {Ziwei Li}, Poulin, Constantinou, {Anirban Sinha}, {Dhruv Balwada}, {SalahKouhen}, Jones, Rocha, Wolfe, {Chuizheng Meng}, Van~Kemenade, Bourbeau, Penn, Busecke, Bueti, and {, Tobias}]{pyqg}
R.~Abernathey, {Rochanotes}, A.~Ross, M.~Jansen, {Ziwei Li}, F.~J. Poulin, N.~C. Constantinou, {Anirban Sinha}, {Dhruv Balwada}, {SalahKouhen}, S.~Jones, C.~B. Rocha, C.~L.~P. Wolfe, {Chuizheng Meng}, H.~Van~Kemenade, J.~Bourbeau, J.~Penn, J.~Busecke, M.~Bueti, and {, Tobias}.
\newblock pyqg/pyqg: v0.7.2, 2022.
\newblock URL \url{https://zenodo.org/record/6563667}.

\bibitem[Arakawa and Lamb(1977)]{ARAKAWA1977}
A.~Arakawa and V.~R. Lamb.
\newblock \emph{Computational Design of the Basic Dynamical Processes of the UCLA General Circulation Model}, page 173–265.
\newblock Elsevier, 1977.
\newblock \doi{10.1016/b978-0-12-460817-7.50009-4}.
\newblock URL \url{http://dx.doi.org/10.1016/B978-0-12-460817-7.50009-4}.

\bibitem[Bracco et~al.(2024)Bracco, Brajard, Dijkstra, Hassanzadeh, Lessig, and Monteleoni]{Bracco2024}
A.~Bracco, J.~Brajard, H.~A. Dijkstra, P.~Hassanzadeh, C.~Lessig, and C.~Monteleoni.
\newblock Machine learning for the physics of climate.
\newblock \emph{Nature Reviews Physics}, 7\penalty0 (1):\penalty0 6–20, Nov. 2024.
\newblock ISSN 2522-5820.
\newblock \doi{10.1038/s42254-024-00776-3}.
\newblock URL \url{http://dx.doi.org/10.1038/s42254-024-00776-3}.

\bibitem[Bracco et~al.(2025)Bracco, Brajard, Dijkstra, Hassanzadeh, Lessig, and Monteleoni]{bracco2025machine}
A.~Bracco, J.~Brajard, H.~A. Dijkstra, P.~Hassanzadeh, C.~Lessig, and C.~Monteleoni.
\newblock Machine learning for the physics of climate.
\newblock \emph{Nature Reviews Physics}, 7\penalty0 (1):\penalty0 6--20, 2025.

\bibitem[Carati et~al.(1995)Carati, Ghosal, and Moin]{Carati1995}
D.~Carati, S.~Ghosal, and P.~Moin.
\newblock On the representation of backscatter in dynamic localization models.
\newblock \emph{Physics of Fluids}, 7\penalty0 (3):\penalty0 606–616, Mar. 1995.
\newblock ISSN 1089-7666.
\newblock \doi{10.1063/1.868585}.
\newblock URL \url{http://dx.doi.org/10.1063/1.868585}.

\bibitem[Chattopadhyay and Hassanzadeh(2023)]{chattopadhyay2023long}
A.~Chattopadhyay and P.~Hassanzadeh.
\newblock Long-term instabilities of deep learning-based digital twins of the climate system: The cause and a solution.
\newblock \emph{arXiv preprint arXiv:2304.07029}, 2023.

\bibitem[Chattopadhyay et~al.(2020)Chattopadhyay, Subel, and Hassanzadeh]{chattopadhyay2020data}
A.~Chattopadhyay, A.~Subel, and P.~Hassanzadeh.
\newblock Data-driven super-parameterization using deep learning: Experimentation with multiscale {L}orenz 96 systems and transfer learning.
\newblock \emph{Journal of Advances in Modeling Earth Systems}, 12\penalty0 (11):\penalty0 e2020MS002084, 2020.

\bibitem[Chattopadhyay et~al.(2024)Chattopadhyay, Gray, Wu, Lowe, and He]{chattopadhyay2024oceannet}
A.~Chattopadhyay, M.~Gray, T.~Wu, A.~B. Lowe, and R.~He.
\newblock Oceannet: A principled neural operator-based digital twin for regional oceans.
\newblock \emph{Scientific Reports}, 14\penalty0 (1):\penalty0 21181, 2024.

\bibitem[Chen et~al.(2021)Chen, Gong, Wan, Deng, Wan, Liu, Chen, and Liu]{Chen2021}
X.~Chen, C.~Gong, Q.~Wan, L.~Deng, Y.~Wan, Y.~Liu, B.~Chen, and J.~Liu.
\newblock Transfer learning for deep neural network-based partial differential equations solving.
\newblock \emph{Advances in Aerodynamics}, 3\penalty0 (1), Dec. 2021.
\newblock ISSN 2524-6992.
\newblock \doi{10.1186/s42774-021-00094-7}.
\newblock URL \url{http://dx.doi.org/10.1186/s42774-021-00094-7}.

\bibitem[Desai et~al.(2021)Desai, Mattheakis, Joy, Protopapas, and Roberts]{Desai2021}
S.~Desai, M.~Mattheakis, H.~Joy, P.~Protopapas, and S.~Roberts.
\newblock One-shot transfer learning of physics-informed neural networks, 2021.
\newblock URL \url{https://arxiv.org/abs/2110.11286}.

\bibitem[Dipankar et~al.(2015)Dipankar, Stevens, Heinze, Moseley, Z\"{a}ngl, Giorgetta, and Brdar]{Dipankar2015}
A.~Dipankar, B.~Stevens, R.~Heinze, C.~Moseley, G.~Z\"{a}ngl, M.~Giorgetta, and S.~Brdar.
\newblock Large eddy simulation using the general circulation model <scp>icon</scp>.
\newblock \emph{Journal of Advances in Modeling Earth Systems}, 7\penalty0 (3):\penalty0 963–986, July 2015.
\newblock ISSN 1942-2466.
\newblock \doi{10.1002/2015ms000431}.
\newblock URL \url{http://dx.doi.org/10.1002/2015MS000431}.

\bibitem[Domaradzki et~al.(1987)Domaradzki, Metcalfe, Rogallo, and Riley]{Domaradzki1987}
J.~A. Domaradzki, R.~W. Metcalfe, R.~S. Rogallo, and J.~J. Riley.
\newblock Analysis of subgrid-scale eddy viscosity with use of results from direct numerical simulations.
\newblock \emph{Physical Review Letters}, 58\penalty0 (6):\penalty0 547–550, Feb. 1987.
\newblock ISSN 0031-9007.
\newblock \doi{10.1103/physrevlett.58.547}.
\newblock URL \url{http://dx.doi.org/10.1103/PhysRevLett.58.547}.

\bibitem[Fox and Orszag(1973)]{fox1973pseudospectral}
D.~G. Fox and S.~A. Orszag.
\newblock Pseudospectral approximation to two-dimensional turbulence.
\newblock \emph{Journal of Computational Physics}, 11\penalty0 (4):\penalty0 612--619, 1973.

\bibitem[Fox(2012)]{Fox2012}
R.~O. Fox.
\newblock Large-eddy-simulation tools for multiphase flows.
\newblock \emph{Annual Review of Fluid Mechanics}, 44\penalty0 (1):\penalty0 47–76, Jan. 2012.
\newblock ISSN 1545-4479.
\newblock \doi{10.1146/annurev-fluid-120710-101118}.
\newblock URL \url{http://dx.doi.org/10.1146/annurev-fluid-120710-101118}.

\bibitem[Fox-Kemper and Menemenlis(2008)]{FoxKemper2008}
B.~Fox-Kemper and D.~Menemenlis.
\newblock \emph{Can large eddy simulation techniques improve mesoscale rich ocean models?}, page 319–337.
\newblock American Geophysical Union, 2008.
\newblock \doi{10.1029/177gm19}.
\newblock URL \url{http://dx.doi.org/10.1029/177GM19}.

\bibitem[Gallet and Ferrari(2021)]{gallet2021quantitative}
B.~Gallet and R.~Ferrari.
\newblock A quantitative scaling theory for meridional heat transport in planetary atmospheres and oceans.
\newblock \emph{AGU Advances}, 2\penalty0 (3):\penalty0 e2020AV000362, 2021.

\bibitem[Gao et~al.(2022)Gao, Cheung, and Ng]{Gao2022}
Y.~Gao, K.~C. Cheung, and M.~K. Ng.
\newblock Svd-pinns: Transfer learning of physics-informed neural networks via singular value decomposition.
\newblock In \emph{2022 IEEE Symposium Series on Computational Intelligence (SSCI)}. IEEE, Dec. 2022.
\newblock \doi{10.1109/ssci51031.2022.10022281}.
\newblock URL \url{http://dx.doi.org/10.1109/SSCI51031.2022.10022281}.

\bibitem[Germano et~al.(1991)Germano, Piomelli, Moin, and Cabot]{Germano1991}
M.~Germano, U.~Piomelli, P.~Moin, and W.~H. Cabot.
\newblock A dynamic subgrid-scale eddy viscosity model.
\newblock \emph{Physics of Fluids A: Fluid Dynamics}, 3\penalty0 (7):\penalty0 1760–1765, July 1991.
\newblock ISSN 0899-8213.
\newblock \doi{10.1063/1.857955}.
\newblock URL \url{http://dx.doi.org/10.1063/1.857955}.

\bibitem[Goswami et~al.(2020)Goswami, Anitescu, Chakraborty, and Rabczuk]{Goswami2020}
S.~Goswami, C.~Anitescu, S.~Chakraborty, and T.~Rabczuk.
\newblock Transfer learning enhanced physics informed neural network for phase-field modeling of fracture.
\newblock \emph{Theoretical and Applied Fracture Mechanics}, 106:\penalty0 102447, Apr. 2020.
\newblock ISSN 0167-8442.
\newblock \doi{10.1016/j.tafmec.2019.102447}.
\newblock URL \url{http://dx.doi.org/10.1016/j.tafmec.2019.102447}.

\bibitem[Goswami et~al.(2022)Goswami, Kontolati, Shields, and Karniadakis]{Goswami2022}
S.~Goswami, K.~Kontolati, M.~D. Shields, and G.~E. Karniadakis.
\newblock Deep transfer operator learning for partial differential equations under conditional shift.
\newblock \emph{Nature Machine Intelligence}, 4\penalty0 (12):\penalty0 1155–1164, Dec. 2022.
\newblock ISSN 2522-5839.
\newblock \doi{10.1038/s42256-022-00569-2}.
\newblock URL \url{http://dx.doi.org/10.1038/s42256-022-00569-2}.

\bibitem[Gray et~al.(2024)Gray, Chattopadhyay, Wu, Lowe, and He]{gray2024long}
M.~A. Gray, A.~Chattopadhyay, T.~Wu, A.~Lowe, and R.~He.
\newblock Long-term prediction of the gulf stream meander using oceannet: a principled neural operator-based digital twin.
\newblock \emph{EGUsphere}, 2024:\penalty0 1--23, 2024.

\bibitem[Grooms(2023{\natexlab{a}})]{Grooms2023}
I.~Grooms.
\newblock Backscatter in energetically-constrained leith parameterizations.
\newblock \emph{Ocean Modelling}, 186:\penalty0 102265, Dec. 2023{\natexlab{a}}.
\newblock ISSN 1463-5003.
\newblock \doi{10.1016/j.ocemod.2023.102265}.
\newblock URL \url{http://dx.doi.org/10.1016/j.ocemod.2023.102265}.

\bibitem[Grooms(2023{\natexlab{b}})]{grooms2023backscatter}
I.~Grooms.
\newblock Backscatter in energetically-constrained leith parameterizations.
\newblock \emph{Ocean Modelling}, 186:\penalty0 102265, 2023{\natexlab{b}}.

\bibitem[Guan et~al.(2022a)Guan, Chattopadhyay, Subel, and Hassanzadeh]{guan2022stable}
Y.~Guan, A.~Chattopadhyay, A.~Subel, and P.~Hassanzadeh.
\newblock Stable a posteriori {LES} of {2D} turbulence using convolutional neural networks: {B}ackscattering analysis and generalization to higher {R}e via transfer learning.
\newblock \emph{Journal of Computational Physics}, 458:\penalty0 111090, 2022a.

\bibitem[Guan et~al.(2023)Guan, Subel, Chattopadhyay, and Hassanzadeh]{guan2022learning}
Y.~Guan, A.~Subel, A.~Chattopadhyay, and P.~Hassanzadeh.
\newblock Learning physics-constrained subgrid-scale closures in the small-data regime for stable and accurate {LES}.
\newblock \emph{Physica D: Nonlinear Phenomena}, 443:\penalty0 133568, 2023.

\bibitem[Guan et~al.(2024)Guan, Hassanzadeh, Schneider, Dunbar, Huang, Wu, and Lopez-Gomez]{Guan2024}
Y.~Guan, P.~Hassanzadeh, T.~Schneider, O.~Dunbar, D.~Z. Huang, J.~Wu, and I.~Lopez-Gomez.
\newblock Online learning of eddy-viscosity and backscattering closures for geophysical turbulence using ensemble kalman inversion, 2024.
\newblock URL \url{https://arxiv.org/abs/2409.04985}.

\bibitem[Guo et~al.(2022)Guo, Zhuang, Chen, Alajlan, and Rabczuk]{Guo2022}
H.~Guo, X.~Zhuang, P.~Chen, N.~Alajlan, and T.~Rabczuk.
\newblock Analysis of three-dimensional potential problems in non-homogeneous media with physics-informed deep collocation method using material transfer learning and sensitivity analysis.
\newblock \emph{Engineering with Computers}, 38\penalty0 (6):\penalty0 5423–5444, Mar. 2022.
\newblock ISSN 1435-5663.
\newblock \doi{10.1007/s00366-022-01633-6}.
\newblock URL \url{http://dx.doi.org/10.1007/s00366-022-01633-6}.

\bibitem[Haghighat et~al.(2021)Haghighat, Raissi, Moure, Gomez, and Juanes]{Haghighat2021}
E.~Haghighat, M.~Raissi, A.~Moure, H.~Gomez, and R.~Juanes.
\newblock A physics-informed deep learning framework for inversion and surrogate modeling in solid mechanics.
\newblock \emph{Computer Methods in Applied Mechanics and Engineering}, 379:\penalty0 113741, June 2021.
\newblock ISSN 0045-7825.
\newblock \doi{10.1016/j.cma.2021.113741}.
\newblock URL \url{http://dx.doi.org/10.1016/j.cma.2021.113741}.

\bibitem[Hallberg(2013)]{hallberg2013using}
R.~Hallberg.
\newblock Using a resolution function to regulate parameterizations of oceanic mesoscale eddy effects.
\newblock \emph{Ocean Modelling}, 72:\penalty0 92--103, 2013.

\bibitem[Hanna et~al.(2022)Hanna, Aguado, Comas-Cardona, Askri, and Borzacchiello]{Hanna2022}
J.~M. Hanna, J.~V. Aguado, S.~Comas-Cardona, R.~Askri, and D.~Borzacchiello.
\newblock Residual-based adaptivity for two-phase flow simulation in porous media using physics-informed neural networks.
\newblock \emph{Computer Methods in Applied Mechanics and Engineering}, 396:\penalty0 115100, June 2022.
\newblock ISSN 0045-7825.
\newblock \doi{10.1016/j.cma.2022.115100}.
\newblock URL \url{http://dx.doi.org/10.1016/j.cma.2022.115100}.

\bibitem[Hornik et~al.(1989)Hornik, Stinchcombe, and White]{Hornik1989}
K.~Hornik, M.~Stinchcombe, and H.~White.
\newblock Multilayer feedforward networks are universal approximators.
\newblock \emph{Neural Networks}, 2\penalty0 (5):\penalty0 359–366, Jan. 1989.
\newblock ISSN 0893-6080.
\newblock \doi{10.1016/0893-6080(89)90020-8}.
\newblock URL \url{http://dx.doi.org/10.1016/0893-6080(89)90020-8}.

\bibitem[Jakhar et~al.(2023)Jakhar, Guan, Mojgani, Chattopadhyay, and Hassanzadeh]{karan2024eqdisc}
K.~Jakhar, Y.~Guan, R.~Mojgani, A.~Chattopadhyay, and P.~Hassanzadeh.
\newblock Learning closed-form equations for subgrid-scale closures from high-fidelity data: Promises and challenges, 2023.
\newblock URL \url{https://arxiv.org/abs/2306.05014}.

\bibitem[Jansen and Held(2014)]{Jansen2014}
M.~F. Jansen and I.~M. Held.
\newblock Parameterizing subgrid-scale eddy effects using energetically consistent backscatter.
\newblock \emph{Ocean Modelling}, 80:\penalty0 36–48, Aug. 2014.
\newblock ISSN 1463-5003.
\newblock \doi{10.1016/j.ocemod.2014.06.002}.
\newblock URL \url{http://dx.doi.org/10.1016/j.ocemod.2014.06.002}.

\bibitem[Jansen et~al.(2019)Jansen, Adcroft, Khani, and Kong]{Jansen2019}
M.~F. Jansen, A.~Adcroft, S.~Khani, and H.~Kong.
\newblock Toward an energetically consistent, resolution aware parameterization of ocean mesoscale eddies.
\newblock \emph{Journal of Advances in Modeling Earth Systems}, 11\penalty0 (8):\penalty0 2844–2860, Aug. 2019.
\newblock ISSN 1942-2466.
\newblock \doi{10.1029/2019ms001750}.
\newblock URL \url{http://dx.doi.org/10.1029/2019MS001750}.

\bibitem[Kent et~al.(2016)Kent, Jablonowski, Thuburn, and Wood]{Kent2016}
J.~Kent, C.~Jablonowski, J.~Thuburn, and N.~Wood.
\newblock An energy‐conserving restoration scheme for the shallow‐water equations.
\newblock \emph{Quarterly Journal of the Royal Meteorological Society}, 142\penalty0 (695):\penalty0 1100–1110, Jan. 2016.
\newblock ISSN 1477-870X.
\newblock \doi{10.1002/qj.2713}.
\newblock URL \url{http://dx.doi.org/10.1002/qj.2713}.

\bibitem[Kerr et~al.(1996)Kerr, Domaradzki, and Barbier]{Kerr1996}
R.~M. Kerr, J.~A. Domaradzki, and G.~Barbier.
\newblock Small-scale properties of nonlinear interactions and subgrid-scale energy transfer in isotropic turbulence.
\newblock \emph{Physics of Fluids}, 8\penalty0 (1):\penalty0 197–208, Jan. 1996.
\newblock ISSN 1089-7666.
\newblock \doi{10.1063/1.868827}.
\newblock URL \url{http://dx.doi.org/10.1063/1.868827}.

\bibitem[Khani and Waite(2016)]{Khani2016}
S.~Khani and M.~L. Waite.
\newblock Backscatter in stratified turbulence.
\newblock \emph{European Journal of Mechanics - B/Fluids}, 60:\penalty0 1–12, Nov. 2016.
\newblock ISSN 0997-7546.
\newblock \doi{10.1016/j.euromechflu.2016.06.012}.
\newblock URL \url{http://dx.doi.org/10.1016/j.euromechflu.2016.06.012}.

\bibitem[Kingma and Ba(2014)]{kingma2014adam}
D.~P. Kingma and J.~Ba.
\newblock Adam: A method for stochastic optimization.
\newblock \emph{arXiv preprint arXiv:1412.6980}, 2014.

\bibitem[Knaepen and Moin(2004)]{Knaepen2004}
B.~Knaepen and P.~Moin.
\newblock Large-eddy simulation of conductive flows at low magnetic reynolds number.
\newblock \emph{Physics of Fluids}, 16\penalty0 (5):\penalty0 1255–1261, May 2004.
\newblock ISSN 1089-7666.
\newblock \doi{10.1063/1.1651484}.
\newblock URL \url{http://dx.doi.org/10.1063/1.1651484}.

\bibitem[Krueger et~al.(2020)Krueger, Caballero, Jacobsen, Zhang, Binas, Zhang, Priol, and Courville]{Krueger2020}
D.~Krueger, E.~Caballero, J.-H. Jacobsen, A.~Zhang, J.~Binas, D.~Zhang, R.~L. Priol, and A.~Courville.
\newblock Out-of-distribution generalization via risk extrapolation (rex), 2020.
\newblock URL \url{https://arxiv.org/abs/2003.00688}.

\bibitem[Larraondo et~al.(2019)Larraondo, Renzullo, Inza, and Lozano]{Larraondo2019}
P.~R. Larraondo, L.~J. Renzullo, I.~Inza, and J.~A. Lozano.
\newblock A data-driven approach to precipitation parameterizations using convolutional encoder-decoder neural networks, 2019.
\newblock URL \url{https://arxiv.org/abs/1903.10274}.

\bibitem[Leslie and Quarini(1979)]{Leslie1979}
D.~C. Leslie and G.~L. Quarini.
\newblock The application of turbulence theory to the formulation of subgrid modelling procedures.
\newblock \emph{Journal of Fluid Mechanics}, 91\penalty0 (01):\penalty0 65, Mar. 1979.
\newblock ISSN 1469-7645.
\newblock \doi{10.1017/s0022112079000045}.
\newblock URL \url{http://dx.doi.org/10.1017/S0022112079000045}.

\bibitem[Li et~al.(2021)Li, Zheng, Kovachki, Jin, Chen, Liu, Azizzadenesheli, and Anandkumar]{Li2021}
Z.~Li, H.~Zheng, N.~Kovachki, D.~Jin, H.~Chen, B.~Liu, K.~Azizzadenesheli, and A.~Anandkumar.
\newblock Physics-informed neural operator for learning partial differential equations, 2021.
\newblock URL \url{https://arxiv.org/abs/2111.03794}.

\bibitem[Lund(1997)]{lund1997use}
T.~Lund.
\newblock On the use of discrete filters for large eddy simulation.
\newblock \emph{Annual Research Briefs}, pages 83--95, 1997.

\bibitem[Lupin-Jimenez et~al.(2025)Lupin-Jimenez, Darman, Hazarika, Wu, Gray, He, Wong, and Chattopadhyay]{lupin2025simultaneous}
L.~Lupin-Jimenez, M.~Darman, S.~Hazarika, T.~Wu, M.~Gray, R.~He, A.~Wong, and A.~Chattopadhyay.
\newblock Simultaneous emulation and downscaling with physically-consistent deep learning-based regional ocean emulators.
\newblock \emph{arXiv preprint arXiv:2501.05058}, 2025.

\bibitem[Mason and Thomson(1992)]{Mason1992}
P.~J. Mason and D.~J. Thomson.
\newblock Stochastic backscatter in large-eddy simulations of boundary layers.
\newblock \emph{Journal of Fluid Mechanics}, 242:\penalty0 51–78, Sept. 1992.
\newblock ISSN 1469-7645.
\newblock \doi{10.1017/s0022112092002271}.
\newblock URL \url{http://dx.doi.org/10.1017/S0022112092002271}.

\bibitem[Mathieu et~al.(2013)Mathieu, Henaff, and LeCun]{mathieu2013fast}
M.~Mathieu, M.~Henaff, and Y.~LeCun.
\newblock Fast training of convolutional networks through ffts.
\newblock \emph{arXiv preprint arXiv:1312.5851}, 2013.

\bibitem[Mattheakis et~al.(2021)Mattheakis, Joy, and Protopapas]{Mattheakis2021}
M.~Mattheakis, H.~Joy, and P.~Protopapas.
\newblock Unsupervised reservoir computing for solving ordinary differential equations, 2021.
\newblock URL \url{https://arxiv.org/abs/2108.11417}.

\bibitem[Meneveau and Katz(2000)]{Meneveau2000}
C.~Meneveau and J.~Katz.
\newblock Scale-invariance and turbulence models for large-eddy simulation.
\newblock \emph{Annual Review of Fluid Mechanics}, 32\penalty0 (1):\penalty0 1–32, Jan. 2000.
\newblock ISSN 1545-4479.
\newblock \doi{10.1146/annurev.fluid.32.1.1}.
\newblock URL \url{http://dx.doi.org/10.1146/annurev.fluid.32.1.1}.

\bibitem[Orszag(1971)]{orszag1971elimination}
S.~A. Orszag.
\newblock On the elimination of aliasing in finite-difference schemes by filtering high-wavenumber components.
\newblock \emph{Journal of Atmospheric Sciences}, 28\penalty0 (6):\penalty0 1074--1074, 1971.

\bibitem[Pahlavan et~al.(2024{\natexlab{a}})Pahlavan, Hassanzadeh, and Alexander]{Pahlavan2024}
H.~A. Pahlavan, P.~Hassanzadeh, and M.~J. Alexander.
\newblock Explainable offline‐online training of neural networks for parameterizations: A 1d gravity wave‐qbo testbed in the small‐data regime.
\newblock \emph{Geophysical Research Letters}, 51\penalty0 (2), Jan. 2024{\natexlab{a}}.
\newblock ISSN 1944-8007.
\newblock \doi{10.1029/2023gl106324}.
\newblock URL \url{http://dx.doi.org/10.1029/2023GL106324}.

\bibitem[Pahlavan et~al.(2024{\natexlab{b}})Pahlavan, Hassanzadeh, and Alexander]{pahlavan2024explainable}
H.~A. Pahlavan, P.~Hassanzadeh, and M.~J. Alexander.
\newblock Explainable offline-online training of neural networks for parameterizations: A 1d gravity wave-qbo testbed in the small-data regime.
\newblock \emph{Geophysical Research Letters}, 51\penalty0 (2):\penalty0 e2023GL106324, 2024{\natexlab{b}}.

\bibitem[Piomelli(1999)]{Piomelli1999}
U.~Piomelli.
\newblock Large-eddy simulation: achievements and challenges.
\newblock \emph{Progress in Aerospace Sciences}, 35\penalty0 (4):\penalty0 335–362, May 1999.
\newblock ISSN 0376-0421.
\newblock \doi{10.1016/s0376-0421(98)00014-1}.
\newblock URL \url{http://dx.doi.org/10.1016/S0376-0421(98)00014-1}.

\bibitem[Pope(2000)]{pope2000turbulent}
S.~Pope.
\newblock \emph{Turbulent flows}.
\newblock Cambridge university press, 2000.

\bibitem[Porta~Mana and Zanna(2014)]{PortaMana2014}
P.~Porta~Mana and L.~Zanna.
\newblock Toward a stochastic parameterization of ocean mesoscale eddies.
\newblock \emph{Ocean Modelling}, 79:\penalty0 1–20, July 2014.
\newblock ISSN 1463-5003.
\newblock \doi{10.1016/j.ocemod.2014.04.002}.
\newblock URL \url{http://dx.doi.org/10.1016/j.ocemod.2014.04.002}.

\bibitem[Pressel et~al.(2017)Pressel, Mishra, Schneider, Kaul, and Tan]{Pressel2017}
K.~G. Pressel, S.~Mishra, T.~Schneider, C.~M. Kaul, and Z.~Tan.
\newblock Numerics and subgrid‐scale modeling in large eddy simulations of stratocumulus clouds.
\newblock \emph{Journal of Advances in Modeling Earth Systems}, 9\penalty0 (2):\penalty0 1342–1365, June 2017.
\newblock ISSN 1942-2466.
\newblock \doi{10.1002/2016ms000778}.
\newblock URL \url{http://dx.doi.org/10.1002/2016MS000778}.

\bibitem[Rasp et~al.(2018)Rasp, Pritchard, and Gentine]{Rasp2018}
S.~Rasp, M.~S. Pritchard, and P.~Gentine.
\newblock Deep learning to represent subgrid processes in climate models.
\newblock \emph{Proceedings of the National Academy of Sciences}, 115\penalty0 (39):\penalty0 9684–9689, Sept. 2018.
\newblock ISSN 1091-6490.
\newblock \doi{10.1073/pnas.1810286115}.
\newblock URL \url{http://dx.doi.org/10.1073/pnas.1810286115}.

\bibitem[Ross et~al.(2023)Ross, Li, Perezhogin, Fernandez-Granda, and Zanna]{ross2023benchmarking}
A.~Ross, Z.~Li, P.~Perezhogin, C.~Fernandez-Granda, and L.~Zanna.
\newblock Benchmarking of machine learning ocean subgrid parameterizations in an idealized model.
\newblock \emph{Journal of Advances in Modeling Earth Systems}, 15\penalty0 (1):\penalty0 e2022MS003258, 2023.

\bibitem[Sagaut(2005)]{sagaut2005large}
P.~Sagaut.
\newblock \emph{Large eddy simulation for incompressible flows: an introduction}.
\newblock Springer Science \& Business Media, 2005.

\bibitem[Sagaut et~al.(2013)Sagaut, Terracol, and Deck]{sagaut2013multiscale}
P.~Sagaut, M.~Terracol, and S.~Deck.
\newblock \emph{Multiscale and multiresolution approaches in turbulence-LES, DES and Hybrid RANS/LES Methods: Applications and Guidelines}.
\newblock World Scientific, 2013.

\bibitem[Sarlak et~al.(2015)Sarlak, Meneveau, and Sørensen]{Sarlak2015}
H.~Sarlak, C.~Meneveau, and J.~Sørensen.
\newblock Role of subgrid-scale modeling in large eddy simulation of wind turbine wake interactions.
\newblock \emph{Renewable Energy}, 77:\penalty0 386–399, May 2015.
\newblock ISSN 0960-1481.
\newblock \doi{10.1016/j.renene.2014.12.036}.
\newblock URL \url{http://dx.doi.org/10.1016/j.renene.2014.12.036}.

\bibitem[Schneider et~al.(2017)Schneider, Lan, Stuart, and Teixeira]{Schneider2017}
T.~Schneider, S.~Lan, A.~Stuart, and J.~Teixeira.
\newblock Earth system modeling 2.0: A blueprint for models that learn from observations and targeted high‐resolution simulations.
\newblock \emph{Geophysical Research Letters}, 44\penalty0 (24), Dec. 2017.
\newblock ISSN 1944-8007.
\newblock \doi{10.1002/2017gl076101}.
\newblock URL \url{http://dx.doi.org/10.1002/2017GL076101}.

\bibitem[Shevchenko and Berloff(2021)]{Shevchenko2021}
I.~Shevchenko and P.~Berloff.
\newblock On a minimum set of equations for parameterisations in comprehensive ocean circulation models.
\newblock \emph{Ocean Modelling}, 168:\penalty0 101913, Dec. 2021.
\newblock ISSN 1463-5003.
\newblock \doi{10.1016/j.ocemod.2021.101913}.
\newblock URL \url{http://dx.doi.org/10.1016/j.ocemod.2021.101913}.

\bibitem[Shinde(2020)]{Shinde2020}
V.~Shinde.
\newblock Proper orthogonal decomposition assisted subfilter-scale model of turbulence for large eddy simulation.
\newblock \emph{Physical Review Fluids}, 5\penalty0 (1), Jan. 2020.
\newblock ISSN 2469-990X.
\newblock \doi{10.1103/physrevfluids.5.014605}.
\newblock URL \url{http://dx.doi.org/10.1103/PhysRevFluids.5.014605}.

\bibitem[Smagorinsky(1963)]{SMAGORINSKY1963}
Smagorinsky.
\newblock General circulation experiments with the primitive equations: I. the basic experiment*.
\newblock \emph{Monthly Weather Review}, 91\penalty0 (3):\penalty0 99–164, Mar. 1963.
\newblock ISSN 1520-0493.
\newblock \doi{10.1175/1520-0493(1963)091<0099:gcewtp>2.3.co;2}.
\newblock URL \url{http://dx.doi.org/10.1175/1520-0493(1963)091<0099:GCEWTP>2.3.CO;2}.

\bibitem[Stevens et~al.(2018)Stevens, Martínez-Tossas, and Meneveau]{Stevens2018}
R.~J. Stevens, L.~A. Martínez-Tossas, and C.~Meneveau.
\newblock Comparison of wind farm large eddy simulations using actuator disk and actuator line models with wind tunnel experiments.
\newblock \emph{Renewable Energy}, 116:\penalty0 470–478, Feb. 2018.
\newblock ISSN 0960-1481.
\newblock \doi{10.1016/j.renene.2017.08.072}.
\newblock URL \url{http://dx.doi.org/10.1016/j.renene.2017.08.072}.

\bibitem[Subel et~al.(2021)Subel, Chattopadhyay, Guan, and Hassanzadeh]{subel2021data}
A.~Subel, A.~Chattopadhyay, Y.~Guan, and P.~Hassanzadeh.
\newblock Data-driven subgrid-scale modeling of forced {B}urgers turbulence using deep learning with generalization to higher {R}eynolds numbers via transfer learning.
\newblock \emph{Physics of Fluids}, 33\penalty0 (3):\penalty0 031702, 2021.

\bibitem[Subel et~al.(2023)Subel, Guan, Chattopadhyay, and Hassanzadeh]{subel2022explaining}
A.~Subel, Y.~Guan, A.~Chattopadhyay, and P.~Hassanzadeh.
\newblock Explaining the physics of transfer learning in data-driven turbulence modeling.
\newblock \emph{PNAS Nexus}, page pgad015, 2023.

\bibitem[Subramanian et~al.(2023)Subramanian, Harrington, Keutzer, Bhimji, Morozov, Mahoney, and Gholami]{Subramanian2023}
S.~Subramanian, P.~Harrington, K.~Keutzer, W.~Bhimji, D.~Morozov, M.~Mahoney, and A.~Gholami.
\newblock Towards foundation models for scientific machine learning: Characterizing scaling and transfer behavior.
\newblock 2023.
\newblock \doi{10.48550/ARXIV.2306.00258}.
\newblock URL \url{https://arxiv.org/abs/2306.00258}.

\bibitem[Sun et~al.(2023)Sun, Pahlavan, Chattopadhyay, Hassanzadeh, Lubis, Alexander, Gerber, Sheshadri, and Guan]{Sun2023}
Y.~Q. Sun, H.~A. Pahlavan, A.~Chattopadhyay, P.~Hassanzadeh, S.~W. Lubis, M.~J. Alexander, E.~Gerber, A.~Sheshadri, and Y.~Guan.
\newblock Data imbalance, uncertainty quantification, and generalization via transfer learning in data-driven parameterizations: Lessons from the emulation of gravity wave momentum transport in waccm, 2023.
\newblock URL \url{https://arxiv.org/abs/2311.17078}.

\bibitem[Tan et~al.(2017)Tan, Schneider, Teixeira, and Pressel]{Tan2017}
Z.~Tan, T.~Schneider, J.~Teixeira, and K.~G. Pressel.
\newblock Large‐eddy simulation of subtropical cloud‐topped boundary layers: 2. cloud response to climate change.
\newblock \emph{Journal of Advances in Modeling Earth Systems}, 9\penalty0 (1):\penalty0 19–38, Jan. 2017.
\newblock ISSN 1942-2466.
\newblock \doi{10.1002/2016ms000804}.
\newblock URL \url{http://dx.doi.org/10.1002/2016MS000804}.

\bibitem[Thuburn et~al.(2013)Thuburn, Kent, and Wood]{Thuburn2013}
J.~Thuburn, J.~Kent, and N.~Wood.
\newblock Cascades, backscatter and conservation in numerical models of two‐dimensional turbulence.
\newblock \emph{Quarterly Journal of the Royal Meteorological Society}, 140\penalty0 (679):\penalty0 626–638, June 2013.
\newblock ISSN 1477-870X.
\newblock \doi{10.1002/qj.2166}.
\newblock URL \url{http://dx.doi.org/10.1002/qj.2166}.

\bibitem[Xu et~al.(2023)Xu, Cao, Yuan, and Meschke]{Xu2023}
C.~Xu, B.~T. Cao, Y.~Yuan, and G.~Meschke.
\newblock Transfer learning based physics-informed neural networks for solving inverse problems in engineering structures under different loading scenarios.
\newblock \emph{Computer Methods in Applied Mechanics and Engineering}, 405:\penalty0 115852, Feb. 2023.
\newblock ISSN 0045-7825.
\newblock \doi{10.1016/j.cma.2022.115852}.
\newblock URL \url{http://dx.doi.org/10.1016/j.cma.2022.115852}.

\bibitem[Xu et~al.(2022)Xu, Lu, and Wang]{Xu2022}
W.~Xu, Y.~Lu, and L.~Wang.
\newblock Transfer learning enhanced deeponet for long-time prediction of evolution equations, 2022.
\newblock URL \url{https://arxiv.org/abs/2212.04663}.

\bibitem[Yosinski et~al.(2014)Yosinski, Clune, Bengio, and Lipson]{Yosinski2014}
J.~Yosinski, J.~Clune, Y.~Bengio, and H.~Lipson.
\newblock How transferable are features in deep neural networks?
\newblock 2014.
\newblock \doi{10.48550/ARXIV.1411.1792}.
\newblock URL \url{https://arxiv.org/abs/1411.1792}.

\bibitem[Zanna and Bolton(2020)]{Zanna2020}
L.~Zanna and T.~Bolton.
\newblock Data‐driven equation discovery of ocean mesoscale closures.
\newblock \emph{Geophysical Research Letters}, 47\penalty0 (17), Aug. 2020.
\newblock ISSN 1944-8007.
\newblock \doi{10.1029/2020gl088376}.
\newblock URL \url{http://dx.doi.org/10.1029/2020GL088376}.

\bibitem[Zhou(1991)]{Zhou1991}
Y.~Zhou.
\newblock Eddy damping, backscatter, and subgrid stresses in subgrid modeling of turbulence.
\newblock \emph{Physical Review A}, 43\penalty0 (12):\penalty0 7049–7052, June 1991.
\newblock ISSN 1094-1622.
\newblock \doi{10.1103/physreva.43.7049}.
\newblock URL \url{http://dx.doi.org/10.1103/PhysRevA.43.7049}.

\bibitem[Zhuang et~al.(2021)Zhuang, Qi, Duan, Xi, Zhu, Zhu, Xiong, and He]{Zhuang2021}
F.~Zhuang, Z.~Qi, K.~Duan, D.~Xi, Y.~Zhu, H.~Zhu, H.~Xiong, and Q.~He.
\newblock A comprehensive survey on transfer learning.
\newblock \emph{Proceedings of the IEEE}, 109\penalty0 (1):\penalty0 43–76, Jan. 2021.
\newblock ISSN 1558-2256.
\newblock \doi{10.1109/jproc.2020.3004555}.
\newblock URL \url{http://dx.doi.org/10.1109/JPROC.2020.3004555}.

\end{thebibliography}
\end{document}